\documentclass[11pt,twoside]{article}

\usepackage{fullpage}

\usepackage{epsf}
\usepackage{fancyhdr}
\usepackage{graphics}
\usepackage{graphicx}
\usepackage{psfrag}
\usepackage{algorithm}
\usepackage{algorithmic}

\usepackage{color}

\usepackage{amsthm}
\usepackage{amsfonts}
\usepackage{amsmath}
\usepackage{amssymb,bbm}
\usepackage[colorlinks]{hyperref}

\usepackage{url}

\usepackage{nicefrac}

\usepackage{chngpage}

 \usepackage{tabularx}%

\usepackage{enumitem}
\usepackage{booktabs}
\usepackage{pbox}

\usepackage{mathtools}

\usepackage{fullpage}
\usepackage{xcolor}
\usepackage{pdfcolmk}
\usepackage{float}
\usepackage{amsmath}
\usepackage{amsthm}
\usepackage{amssymb}
\usepackage{esint}
\usepackage{xargs}[2008/03/08]
\PassOptionsToPackage{normalem}{ulem}
\usepackage{ulem}
\usepackage{graphicx}
\usepackage{balance}
\usepackage{subfig}
\usepackage{mathtools, nccmath}

\theoremstyle{plain}
\newtheorem{thm}{\protect\theoremname}
\ifx\proof\undefined
\newenvironment{proof}[1][\protect\proofname]{\par
\normalfont\topsep6\p@\@plus6\p@\relax
\trivlist
\itemindent\parindent
\item[\hskip\labelsep\scshape #1]\ignorespaces
}{%
\endtrivlist\@endpefalse
}
\providecommand{\proofname}{Proof}
\fi
\makeatother

\usepackage[numbers]{natbib}
\providecommand{\theoremname}{Theorem}

\numberwithin{theorem}{section}

\numberwithin{proposition}{section}

\numberwithin{lemma}{section}

\numberwithin{definition}{section}

\numberwithin{condition}{section}

\numberwithin{problem}{section}

\numberwithin{corollary}{section}

\numberwithin{assumption}{section}

\numberwithin{example}{section}

\long\def\comment#1{}
\comment{
\setlength{\topmargin}{0 in}
\setlength{\textwidth}{5.5 in}
\setlength{\textheight}{8 in}
\setlength{\oddsidemargin}{0.5 in}
}

\begin{document}

\begin{center}
{\bf{\LARGE{Probabilistic Multilevel Clustering via Composite Transportation Distance}}}

\vspace*{.2in}
 {\large{
 \begin{tabular}{cc}
  Nhat Ho$^{\star, \dagger}$ &  Viet Huynh$^{\star, \ddagger}$ \\
 \end{tabular}
 \begin{tabular}
 {cc}
  Dinh Phung$^{\ddagger}$ & Michael I. Jordan$^{\dagger}$
 \end{tabular}
}}

 \vspace*{.2in}

 \begin{tabular}{c}
University of California at Berkeley, Berkeley, USA $^\dagger$ \\
Monash University, Australia $^\ddagger$
\end{tabular}

\vspace*{.2in}

\today

\vspace*{.2in}
\begin{abstract}

We propose a novel probabilistic approach to multilevel clustering problems based on composite transportation distance, which is a variant of transportation distance where the underlying metric is Kullback-Leibler divergence. Our method involves solving a joint optimization problem over spaces of probability measures to simultaneously discover grouping structures within groups and among groups. By exploiting the connection of our method to the problem of finding composite transportation barycenters, we develop fast and efficient optimization algorithms even for potentially large-scale multilevel datasets. Finally, we present experimental results with both synthetic and real data to demonstrate the efficiency and scalability of the proposed approach.

\newcommand{\sidenote}[1]{\marginpar{\small \emph{\color{Medium}#1}}}

\global\long\def\se{\hat{\text{se}}}

\global\long\def\interior{\text{int}}

\global\long\def\boundary{\text{bd}}

\global\long\def\ML{\textsf{ML}}

\global\long\def\GML{\mathsf{GML}}

\global\long\def\HMM{\mathsf{HMM}}

\global\long\def\support{\text{supp}}

\global\long\def\new{\text{*}}

\global\long\def\stir{\text{Stirl}}

\global\long\def\mA{\mathcal{A}}

\global\long\def\mB{\mathcal{B}}

\global\long\def\mF{\mathcal{F}}

\global\long\def\mK{\mathcal{K}}

\global\long\def\mH{\mathcal{H}}

\global\long\def\mM{\mathcal{M}}

\global\long\def\mX{\mathcal{X}}

\global\long\def\mY{\mathcal{Y}}

\global\long\def\mZ{\mathcal{Z}}

\global\long\def\mS{\mathcal{S}}

\global\long\def\Ical{\mathcal{I}}

\global\long\def\mT{\mathcal{T}}

\global\long\def\mP{\mathcal{P}}

\global\long\def\Pcal{\mathcal{P}}

\global\long\def\dist{d}

\global\long\def\HX{\entro\left(X\right)}
 \global\long\def\entropyX{\HX}

\global\long\def\HY{\entro\left(Y\right)}
 \global\long\def\entropyY{\HY}

\global\long\def\HXY{\entro\left(X,Y\right)}
 \global\long\def\entropyXY{\HXY}

\global\long\def\mutualXY{\mutual\left(X;Y\right)}
 \global\long\def\mutinfoXY{\mutualXY}

\global\long\def\given{\mid}

\global\long\def\gv{\given}

\global\long\def\goto{\rightarrow}

\global\long\def\asgoto{\stackrel{a.s.}{\longrightarrow}}

\global\long\def\pgoto{\stackrel{p}{\longrightarrow}}

\global\long\def\dgoto{\stackrel{d}{\longrightarrow}}

\global\long\def\lik{\mathcal{L}}

\global\long\def\logll{\mathit{l}}

\global\long\def\vectorize#1{\mathbf{#1}}

\global\long\def\vt#1{\mathbf{#1}}

\global\long\def\gvt#1{\boldsymbol{#1}}

\global\long\def\idp{\ \bot\negthickspace\negthickspace\bot\ }
 \global\long\def\cdp{\idp}

\global\long\def\das{:=}

\global\long\def\id{\mathbb{I}}

\global\long\def\idarg#1#2{\id\left\{  #1,#2\right\}  }

\global\long\def\iid{\stackrel{\text{iid}}{\sim}}

\global\long\def\bzero{\vt 0}

\global\long\def\bone{\mathbf{1}}

\global\long\def\boldm{\boldsymbol{m}}

\global\long\def\bff{\vt f}

\global\long\def\bx{\boldsymbol{x}}

\global\long\def\ba{\boldsymbol{a}}

\global\long\def\bb{\boldsymbol{b}}

\global\long\def\bc{\boldsymbol{c}}

\global\long\def\bd{\boldsymbol{d}}

\global\long\def\bl{\boldsymbol{l}}

\global\long\def\bu{\boldsymbol{u}}

\global\long\def\bo{\boldsymbol{o}}

\global\long\def\bh{\boldsymbol{h}}

\global\long\def\bH{\boldsymbol{H}}

\global\long\def\bs{\boldsymbol{s}}

\global\long\def\bz{\boldsymbol{z}}

\global\long\def\xnew{y}

\global\long\def\bxnew{\boldsymbol{y}}

\global\long\def\bX{\boldsymbol{X}}

\global\long\def\tbx{\tilde{\bx}}

\global\long\def\by{\boldsymbol{y}}

\global\long\def\bY{\boldsymbol{Y}}

\global\long\def\bZ{\boldsymbol{Z}}

\global\long\def\bU{\boldsymbol{U}}

\global\long\def\bv{\boldsymbol{v}}

\global\long\def\bn{\boldsymbol{n}}

\global\long\def\bV{\boldsymbol{V}}

\global\long\def\bI{\boldsymbol{I}}

\global\long\def\bw{\vt w}

\global\long\def\bW{\boldsymbol{W}}

\global\long\def\balpha{\gvt{\alpha}}

\global\long\def\bbeta{\gvt{\beta}}

\global\long\def\bmu{\gvt{\mu}}

\global\long\def\btheta{\boldsymbol{\theta}}

\global\long\def\blambda{\boldsymbol{\lambda}}

\global\long\def\bpsi{\boldsymbol{\psi}}

\global\long\def\bphi{\boldsymbol{\phi}}

\global\long\def\bpi{\boldsymbol{\pi}}

\global\long\def\bomega{\boldsymbol{\omega}}

\global\long\def\bepsilon{\boldsymbol{\epsilon}}

\global\long\def\btau{\boldsymbol{\tau}}

\global\long\def\bxi{\boldsymbol{\xi}}

\global\long\def\realset{\mathbb{R}}

\global\long\def\realn{\realset^{n}}

\global\long\def\reald{\realset^{d}}

\global\long\def\integerset{\mathbb{Z}}

\global\long\def\natset{\integerset}

\global\long\def\integer{\integerset}

\global\long\def\natn{\natset^{n}}

\global\long\def\rational{\mathbb{Q}}

\global\long\def\rationaln{\rational^{n}}

\global\long\def\complexset{\mathbb{C}}

\global\long\def\comp{\complexset}

\global\long\def\compl#1{#1^{\text{c}}}

\global\long\def\and{\cap}

\global\long\def\compn{\comp^{n}}

\global\long\def\comb#1#2{\left({#1\atop #2}\right) }

\global\long\def\nchoosek#1#2{\left({#1\atop #2}\right)}

\global\long\def\param{\vt w}

\global\long\def\Param{\Theta}

\global\long\def\meanparam{\gvt{\mu}}

\global\long\def\Meanparam{\mathcal{M}}

\global\long\def\meanmap{\mathbf{m}}

\global\long\def\logpart{A}

\global\long\def\simplex{\Delta}

\global\long\def\simplexn{\simplex^{n}}

\global\long\def\dirproc{\text{DP}}

\global\long\def\ggproc{\text{GG}}

\global\long\def\DP{\text{DP}}

\global\long\def\ndp{\text{nDP}}

\global\long\def\hdp{\text{HDP}}

\global\long\def\gempdf{\text{GEM}}

\global\long\def\rfs{\text{RFS}}

\global\long\def\bernrfs{\text{BernoulliRFS}}

\global\long\def\poissrfs{\text{PoissonRFS}}

\global\long\def\iidrfs{$\text{IidRFS}$}

\global\long\def\grad{\gradient}
 \global\long\def\gradient{\nabla}

\global\long\def\partdev#1#2{\partialdev{#1}{#2}}
 \global\long\def\partialdev#1#2{\frac{\partial#1}{\partial#2}}

\global\long\def\partddev#1#2{\partialdevdev{#1}{#2}}
 \global\long\def\partialdevdev#1#2{\frac{\partial^{2}#1}{\partial#2\partial#2^{\top}}}

\global\long\def\hessian{\text{Hess}}

\global\long\def\closure{\text{cl}}

\global\long\def\cpr#1#2{\Pr\left(#1\ |\ #2\right)}

\global\long\def\var{\text{Var}}

\global\long\def\Var#1{\text{Var}\left[#1\right]}

\global\long\def\cov{\text{Cov}}

\global\long\def\Cov#1{\cov\left[ #1 \right]}

\global\long\def\COV#1#2{\underset{#2}{\cov}\left[ #1 \right]}

\global\long\def\corr{\text{Corr}}

\global\long\def\sst{\text{T}}

\global\long\def\SST{\sst}

\global\long\def\ess{\mathbb{E}}

\global\long\def\Ess#1{\mathbb{E}\left[#1\right]}

\newcommandx\ESS[2][usedefault, addprefix=\global, 1=]{\underset{#2}{\mathbb{E}}\left[#1\right]}

\global\long\def\fisher{\mathcal{F}}

\global\long\def\bfield{\mathcal{B}}
 \global\long\def\borel{\mathcal{B}}

\global\long\def\bernpdf{\text{Bernoulli}}

\global\long\def\betapdf{\text{Beta}}

\global\long\def\dirpdf{\text{Dir}}

\global\long\def\gammapdf{\text{Gamma}}

\global\long\def\gaussden#1#2{\text{Normal}\left(#1, #2 \right) }

\global\long\def\gauss{\mathbf{N}}

\global\long\def\gausspdf#1#2#3{\text{Normal}\left( #1 \lcabra{#2, #3}\right) }

\global\long\def\multpdf{\text{Mult}}

\global\long\def\poiss{\text{Pois}}

\global\long\def\poissonpdf{\text{Poisson}}

\global\long\def\pgpdf{\text{PG}}

\global\long\def\wshpdf{\text{Wish}}

\global\long\def\iwshpdf{\text{InvWish}}

\global\long\def\nwpdf{\text{NW}}

\global\long\def\niwpdf{\text{NIW}}

\global\long\def\studentpdf{\text{Student}}

\global\long\def\unipdf{\text{Uni}}

\global\long\def\transp#1{\transpose{#1}}
 \global\long\def\transpose#1{#1^{\mathsf{T}}}

\global\long\def\mgt{\succ}

\global\long\def\mge{\succeq}

\global\long\def\idenmat{\mathbf{I}}

\global\long\def\trace{\mathrm{tr}}

\global\long\def\argmax#1{\underset{_{#1}}{\text{argmax}} }

\global\long\def\argmin#1{\underset{_{#1}}{\text{argmin}\ } }

\global\long\def\diag{\text{diag}}

\global\long\def\norm{}

\global\long\def\spn{\text{span}}

\global\long\def\vtspace{\mathcal{V}}

\global\long\def\field{\mathcal{F}}
 \global\long\def\ffield{\mathcal{F}}

\global\long\def\inner#1#2{\left\langle #1,#2\right\rangle }
 \global\long\def\iprod#1#2{\inner{#1}{#2}}

\global\long\def\dprod#1#2{#1 \cdot#2}

\global\long\def\norm#1{\left\Vert #1\right\Vert }

\global\long\def\entro{\mathbb{H}}

\global\long\def\entropy{\mathbb{H}}

\global\long\def\Entro#1{\entro\left[#1\right]}

\global\long\def\Entropy#1{\Entro{#1}}

\global\long\def\mutinfo{\mathbb{I}}

\global\long\def\relH{\mathit{D}}

\global\long\def\reldiv#1#2{\relH\left(#1||#2\right)}

\global\long\def\KL{\text{KL}}

\global\long\def\JS{JS}

\global\long\def\KLdiv#1#2{\KL\left(#1\parallel#2\right)}
 \global\long\def\KLdivergence#1#2{\KL\left(#1\ \parallel\ #2\right)}

\global\long\def\JSdiv#1#2{\JS\left(#1\parallel#2\right)}
 \global\long\def\JSdivergence#1#2{\JS\left(#1\ \parallel\ #2\right)}

\global\long\def\crossH{\mathcal{C}}
 \global\long\def\crossentropy{\mathcal{C}}

\global\long\def\crossHxy#1#2{\crossentropy\left(#1\parallel#2\right)}

\global\long\def\breg{\text{BD}}

\global\long\def\lcabra#1{\left|#1\right.}

\global\long\def\lbra#1{\lcabra{#1}}

\global\long\def\rcabra#1{\left.#1\right|}

\global\long\def\rbra#1{\rcabra{#1}}


\global\long\def\compoop{\widehat{W}}
\global\long\def\Ocal{\mathcal{O}}
\global\long\def\bTheta{\boldsymbol{\Theta}}
\global\long\def\bgamma{\boldsymbol{M}}
\global\long\def\prospace{\mathcal{P}}
\global\long\def\glomeas{\boldsymbol{\mathcal{Q}}}
\global\long\def\locmeas{\boldsymbol{P}}
\global\long\def\bPsi{\boldsymbol{\Psi}}
\global\long\def\bglo{\boldsymbol{\gamma}}
\global\long\def\bb{\boldsymbol{b}}
\global\long\def\gloclus{\boldsymbol{C}}

\end{abstract}
\let\thefootnote\relax\footnotetext{$^\star$ Nhat Ho and Viet Huynh contributed equally to this work and are in alphabetical order.}
\end{center}

\section{Introduction}

Clustering is a classic and fundamental problem in machine 
learning. Popular clustering methods such as K-means and the EM algorithm 
have been the workhorses of exploratory data analysis. However, the underlying model for such methods is a simple flat partition or a mixture model, which do not capture multilevel structures (e.g., 
words are grouped into documents, documents are grouped into corpora) that arise in many applications in the physical, biological or cognitive sciences. The clustering of multilevel structured data calls for novel methodologies beyond classical clustering.

One natural approach for capturing  multilevel structures is to use a hierarchy in which data are clustered locally into groups, and those groups are partitioned in a ``global clustering.'' Attempts to develop algorithms of this kind can be roughly classified
into two categories. The first category makes use of probabilistic models, often based on  Dirichlet process priors. Examples in this vein include the Hierarchical Dirichlet Process (HDP)~\cite{Teh-et-al06}, 
Nested Dirichlet Process (NDP)~\cite{Abel-2008}, Multilevel Clustering 
with Context (MC$^{2}$)~\cite{Vu-2014}, and Multilevel Clustering 
Hierarchical Dirichlet Process (MLC-HDP)~\cite{Wulsin-2016}. Despite the 
flexibility and solid statistical foundation of these models, they have seen limited application to large-scale datasets, given concerns about the computational scaling of the sampling-based algorithms that are generally used for  inference under these models.

A second category of multilevel methods is based on  tools from optimal transport theory, where algorithms such as Wasserstein barycenters provide scalable computation~\cite{cuturi2013sinkhorn,Cuturi-2014}. These methods trace their origins to a seminal paper by~\cite{Pollard-1982} which established a connection between the K-means algorithm and the problem of determining a discrete probability measure that is close in Wasserstein distance~\cite{Villani-03} to the empirical measure of the data. Based on this connection, it is possible to use Wasserstein distance to develop a combined local/global multilevel clustering method~\cite{ho17multilevel}.

The specific multilevel clustering method proposed in~\cite{ho17multilevel} has, however, its limitations. Most notably, as that method uses K-means as a building block, it is only applicable to continuous data. When being used to cluster discrete  data, it yields poor results. In this work, we make use of a novel form of transportation distance, which is termed as \textit{composite transportation distance}~\cite{Nguyen-13}, to overcome this limitation, and to provide a more general multilevel clustering method. The salient feature of composite transportation distance is that it utilizes Kullback-Leibler (KL) divergence as the underlying metric of optimal transportation distance, in contrast to the standard Euclidean metric that has been used in optimal transportation approaches to clustering to date.

In order to motivate our use of composite transportation distance, we start with a one- level structure data in which the data  are generated from a finite mixture model, e.g., a mixture of (multivariate) Gaussian distributions or multinomial distributions.  Unlike traditional estimators such as maximum likelihood estimation (MLE), the high-level idea of using composite transportation distance is to determine optimal parameters to minimize the KL cost of moving the likelihood from one cluster to another cluster. Intuitively, with such a  distance, we can employ the underlying geometric structure of parameters to perform efficient clustering with the data. Another advantage of composite transportation distance is its flexibility to generalize to multilevel structure data. More precisely, by representing each group in a multilevel clustering problem by an unknown mixture model (local clustering), we can determine the optimal parameters, which can be represented as local (probability) measures, of each group via optimization problems based on composite transportation distance. Then, in order to determine global clustering among these groups, we perform a composite transportation barycenter problem over the local measures to obtain a global measure over the space of mixture models, which serves as a partition of these groups. As a result, our final method, which we refer to as \textit{multilevel composite transportation} (MCT), involves solving a joint optimal transport optimization problem with respect to both a local clustering and a global clustering based on the cost matrix encoding KL divergence among atoms. The solution strategy involves using the fast computation method of Wasserstein barycenters combined with coordinate descent.

In summary, our main contributions are the following: (i) A new optimization formulation for clustering based on a variety of multilevel data types, including both continuous and discrete observations, based on composite transportation distance; (ii) We provide a highly scalable solution strategy for this optimization formulation; (iii) Although our approach avoids the use of the Dirichlet process as a building block, the approach has much of the flexibility the hierarchical Dirichlet process in its ability to share atoms among local clusterings. We thus are able to borrow strength among clusters, which improves statistical efficiency under certain applications, e.g., image annotation in computer vision.

The paper is organized as follows. Section~\ref{sec:background} provides preliminary background on composite transportation distance and composite transportation barycenters. Section~\ref{sec:model} formulates the multilevel composite transportation optimization model, while Section~\ref{sec:experiments} presents simulation studies with both synthetic and real data. Finally, we conclude the paper with a discussion in Section \ref{sec:conclusion}.  Technical details of proofs and algorithm development are provided in the Appendix.

\section{Composite transportation distance}
\label{sec:background}

Throughout this paper, we let $\Theta$ be a bounded subset of $\mathbb{R}^{d}$ for a given dimension $d \geq 1$. Additionally, $\{f(x|\theta), \ \theta \in \Theta \}$ is a given exponential family of distributions with natural parameter $\theta$:
\begin{align*}
f(x|\theta) : = h(x) \exp\left(\left\langle T\left(x\right),\theta\right\rangle -A\left(\theta\right)\right),
\end{align*}
where $A\left(\theta\right)$ is the log-partition function which is convex. We define $P_{\theta}$ to be the probability distribution whose density function is $f(x|\theta)$. Given a fixed number of $K$ components, we denote a finite mixture distribution as follows:\useshortskip\begin{align}
P_{\bomega_{K},\boldsymbol{\Theta}_{K}} & : =\sum_{k=1}^{K}\omega_{k}P_{\theta_{k}},\label{eq:EF_Prob}
\end{align} where $\bomega_{K} = \left(\omega_{1},\ldots,\omega_{K}\right)\in\simplex^{K}$, which is a probability simplex in $K-1$ dimensions, and $\boldsymbol{\Theta}_{K}=\left\{ \theta_{k}\right\} _{k=1}^{K}\in\Theta^{K}$ are the weights and atoms. Then, the probability density function of mixture model can be expressed\useshortskip\begin{align*}
p_{\bomega_{K},\boldsymbol{\Theta}_{K}}\left(\bx\right) & := \sum_{k=1}^{K}\omega_{k}f(x|\theta_{k}).
\end{align*}
We also use $Q_{\bomega_{K},\bTheta_{K}}$ to denote a finite mixture of at most $K$ components to avoid potential notational clutter.
\comment{For any given subset $\Theta\subset\mathbb{R}^{d}$, let $\mathcal{P}(\Theta)$
denote the space of Borel probability measures on $\Theta$. The Wasserstein
space of order $r\in[1,\infty)$ of probability measures on $\Theta$
is defined as $\mathcal{P}_{r}(\Theta)=\biggr\{ G\in\mathcal{P}(\Theta):\ \int\|x\|^{r}dG(x)<\infty\biggr\}$,
where $\|.\|$ denotes Euclidean metric in $\mathbb{R}^{d}$. Additionally,
for any $k\geq1$ the probability simplex is denoted by $\Delta_{k}=\left\{ u\in\mathbb{R}^{k}:\ u_{i}\geq0,\ \sum\limits _{i=1}^{k}u_{i}=1\right\} $.
Finally, let $\mathcal{O}_{k}(\Theta)$ (resp., $\mathcal{E}_{k}(\Theta)$)
be the set of probability measures with at most (resp., exactly) $k$
support points in $\Theta$.}
\subsection{Composite transportation distance}
\label{Section:composite_transport}  For any two finite mixture probability distributions $P_{\bomega_{K},\boldsymbol{\Theta}_{K}}$ and $P_{\bomega_{K'}',\boldsymbol{\Theta}_{K'}'}$ and any two given numbers $K$ and $K'$, we define the composite transportation distance between $P_{\bomega_{K},\boldsymbol{\Theta}_{K}}$ and $P_{\bomega_{K'}',\boldsymbol{\Theta}_{K'}'}$ as follows\begin{align}
\compoop(P_{\bomega_{K},\boldsymbol{\Theta}_{K}}, P_{\bomega_{K'}',\boldsymbol{\Theta}_{K'}'}) : = \inf_{\bpi\in\Pi\left(\bomega_{K}, \bomega_{K'}' \right)} \left\langle \bpi,\bgamma\right\rangle, \label{eqn:comp_transport}
\end{align}
where the cost matrix $\bgamma = (M_{ij})$ satisfies $M_{ij} = \KL(f(x|\theta_{i}), f(x|\theta_{j}'))$ for $1 \leq i \leq K$ and $1 \leq j \leq K'$. Here, $\left\langle . , . \right \rangle$ denotes the dot product (or Frobenius inner product) of two matrices and $\Pi(\bomega_{K},\bomega_{K'}')$ is the set of all probability measures (or equivalently transportation plans) $\bpi$ on $[0,1]^{K \times K'}$
that have marginals $\bomega_{K}$ and $\bomega_{K'}'$ respectively. 
\paragraph{Detailed form of cost matrix} Since $f(x|\theta)$ is an exponential family, we can compute $\KL\left(f(x|\theta), f(x|\theta')\right)$ in closed form as follows ~\cite[Ch.8]{zhang2009line,jordan2003introduction}\footnote{Note that the order of parameters is reversed in KL and Bregman divergences.}:
\begin{align*}
\KL\left(f(x|\theta'),f(x|\theta)\right) & = D_{A}\left(\theta,\theta'\right),
\end{align*}
where $D_{A}\left(\cdot,\cdot\right)$ is the Bregman divergence associated with log-partition function $A\left(\cdot\right)$ of $f$, i.e.,\begin{align*}
D_{A}\left(\theta,\theta' \right) & = A\left(\theta \right)-A\left(\theta' \right)-\left\langle \grad A\left(\theta' \right),\left(\theta - \theta' \right)\right\rangle .
\end{align*}
Therefore, the cost matrix $\bgamma$ has an explicit form \begin{align}
M_{ij} = A \left(\theta_{i} \right)-A\left(\theta_{j}' \right)-\left\langle \grad A\left(\theta_{j}' \right),\left(\theta_{i} - \theta_{j}' \right)\right\rangle \label{eq:cost_matrix}
\end{align}
for $1 \leq i \leq K$ and $1 \leq j \leq K'$.
\paragraph{Composite transportation distance on the space of finite mixtures of finite mixtures} We can recursively define finite mixtures of finite mixtures, and define a suitable version of composite transportation distance on this abstract space. In particular, consider a collection of $N$ finite mixture probability distributions with at most $K$ components $\left\{P_{\bomega_{K}^{i},\boldsymbol{\Theta}_{K}^{i}}\right\}_{i=1}^{N}$ and a collection of $\bar{N}$ finite mixture probability distributions with at most $\bar{K}$ components $\left\{P_{\overline{\bomega}_{\bar{K}}^{i},\overline{\boldsymbol{\Theta}}_{\bar{K}}^{i}}\right\}_{i=1}^{\bar{N}}$. We define two finite mixtures of these distributions as follows\begin{align}
\mathcal{P} = \sum_{i=1}^{N} \tau_{i}P_{\bomega_{K}^{i},\boldsymbol{\Theta}_{K}^{i}}, \ \mathcal{Q} = \sum_{i=1}^{\bar{N}} \overline{\tau}_{i}P_{\overline{\bomega}_{\bar{K}}^{i},\overline{\boldsymbol{\Theta}}_{\bar{K}}^{i}}, \nonumber 
\end{align}
where $\boldsymbol{\tau} = (\tau_{1}, \ldots, \tau_{N}) \in \Delta^{N}$ and $\boldsymbol{\overline{\tau}} = (\overline{\tau}_{1}, \ldots, \overline{\tau}_{\bar{N}}) \in \Delta^{\bar{N}}$. Then, the composite transportation distance between $\mathcal{P}$ and $\mathcal{Q}$ is\begin{align*}
\compoop(\mathcal{P},\mathcal{Q}) & : = \inf_{\bpi \in \Pi\left(\boldsymbol{\tau}, \boldsymbol{\overline{\tau}} \right)} \left\langle \bpi, \overline{\bgamma} \right\rangle,
\end{align*}
where the cost matrix $\overline{\bgamma} = \left\{\overline{M}_{ij}\right\}$ is defined as
\begin{align*}
\overline{M}_{ij} = \compoop(P_{\bomega_{K}^{i},\boldsymbol{\Theta}_{K}^{i}}, P_{\overline{\bomega}_{\bar{K}}^{j},\overline{\boldsymbol{\Theta}}_{\bar{K}}^{j}}),
\end{align*}
for $1 \leq i \leq N$ and $1 \leq j \leq \bar{N}$. Note that, in a slight notational abuse, $\compoop(.,.)$ is used for both the finite mixtures and finite mixtures of finite mixtures.

\subsection{Learning finite mixtures with composite transportation distance}
In this section, we assume that $X_{1},\ldots,X_{n}$ are i.i.d.\ samples from the mixture density 
\begin{align*}
p_{\bomega_{k_{0}}^{0},\boldsymbol{\Theta}_{k_{0}}^{0}}(x) = \sum_{i=1}^{k_{0}} \omega_{i}^{0}f(x|\theta_{i}^{0}),
\end{align*} 
where $k_{0}< \infty$ is the true number of components. Since $k_{0}$ is generally unknown, we fit this model by a mixture of $K$ distributions where $K \geq k_{0}$.
\subsubsection{Inference with composite transportation distance} 
Denote $P_{n} : = \frac{1}{n} \sum_{i=1}^{n} \delta_{X_{i}}$ as an empirical measure with respect to samples $X_{1},\ldots,X_{n}$. To facilitate the discussion, we define the following composite transportation distance between an empirical measure $P_{n}$ and the mixture probability distribution $P_{\bomega_{K}, \bTheta_{K}}$
\begin{align}
\compoop(P_{n}, P_{\bomega_{K}, \bTheta_{K}}) : = \inf_{\bpi\in\Pi\left(\frac{1}{n}\boldsymbol{1}_{n},\bomega_{K}\right)}\left\langle \bpi,\bgamma\right\rangle, \label{eq:OT_KLcost}
\end{align}
where $\bgamma = (M_{ij}) \in \mathbb{R}^{n \times K}$ is a cost matrix defined as $M_{ij} : = - \log f(X_{i}| \theta_{j})$ for $1 \leq i \leq n, 1 \leq j \leq K$. Furthermore, $\Pi\left(\cdot,\cdot\right)$ is the set of transportation plans between $\boldsymbol{1}_{n}/n$ and $\bomega_{K}$.  

To estimate the true weights $\bomega_{k_{0}}^{0}$ and true components $\theta_{i}^{0}$ as $1 \leq i \leq k_{0}$, we perform an optimization with transportation distance $\compoop$ as follows:
\begin{align}
(\widehat{\bomega}_{n,K}, \widehat{\bTheta}_{n,K}) = \mathop {\arg \min} \limits_{\bomega_{K}, \bTheta_{K}} \compoop(P_{n}, P_{\bomega_{K}, \bTheta_{K}}). \label{eq:optimal_compo}
\end{align}
The estimator $(\widehat{\bomega}_{n,K}, \widehat{\bTheta}_{n,K})$ is usually referred to as the Minimum Kantorovitch estimator~\cite{bassetti2006minimum}.
\begin{algorithm}
\begin{algorithmic} \REQUIRE
Data $D=\left\{ X_{i}\right\} _{i=1}^{n}$; the number of clusters
$K$ the regularized hyper-parameter $\lambda>0$. \ENSURE Optimal
weight-atoms $\left\{ \omega_{j},\theta_{j}\right\} _{j=1}^{K}$
\STATE Initialize weights $\left\{ \omega_{j}\right\} _{j=1}^{K}$
and atoms $\left\{ \theta_{j}\right\} _{j=1}^{K}$. \WHILE{not converged}
\STATE 1. Update weights $\omega_{j}$: \FOR{$j=1$ \TO $K$}
\STATE Compute transportation plan $\pi_{ij}$ as 
\begin{align*}
\pi_{ij}= & \left(f(X_{i}|\theta_{j})\right)^{1/\lambda} \big /\left(n \sum_{k=1}^{K}\left(f(X_{i}|\theta_{j})\right)^{1/\lambda}\right)
\end{align*} for $1\leq i\leq n$\STATE Update weight $\omega_{j}=\sum_{i=1}^{n}\pi_{ij}$.
\ENDFOR \STATE 2. Update atoms $\theta_{j}$: \FOR{$j=1$ \TO
$J$} \STATE Update atoms $\theta_{j}$ as solution of equation $\grad A\left( \theta_{j} \right) = \sum_{i = 1}^{n} \frac{\pi_{ij}}{\omega_{j}} T(X_{i})$.
\ENDFOR \ENDWHILE \end{algorithmic} \caption{\label{algo:local_solution_mixture}Composite Transportation Distance
with Mixtures}
\end{algorithm}

\subsubsection{Regularized composite transportation distance}
As is the case with the traditional optimal transportation distance, the composite transportation distance $\compoop$ does not have a favorable computational complexity. Therefore, we consider an entropic regularizer to speed up its computation~\cite{cuturi2013sinkhorn}. More precisely, we consider the following regularized version of $\compoop(P_{n}, P_{\bomega_{K}, \bTheta_{K}})$:
\begin{align*}
\inf_{\bpi\in\Pi\left(\frac{1}{n}\boldsymbol{1}_{n},\bomega_{K} \right)} \left\langle \bpi,\bgamma\right\rangle - \lambda\entro\left(\bpi\right), \nonumber
\end{align*}
where $\lambda>0$ is a penalization term and $\entro\left(\bpi\right) : = - \sum_{i,j} \pi_{ij}\log \pi_{ij}$ is an entropy of $\bpi \in \Pi\left(\boldsymbol{1}_{n}/n,\bomega_{K} \right)$. Equipped with this regularization, we have a regularized version of the optimal estimator in~\eqref{eq:optimal_compo}:
\begin{align}
\min \limits_{\bomega_{K}, \bTheta_{K}} \inf_{\bpi\in\Pi\left(\frac{1}{n}\boldsymbol{1}_{n},\bomega_{K} \right)} \left\langle \bpi,\bgamma\right\rangle - \lambda\entro\left(\bpi\right). \label{eq:regu_optimal_compo}
\end{align}
We summarize the algorithm for determining local solutions of the above objective function in Algorithm~\ref{algo:local_solution_mixture}. The details for how to obtain the updates of weight and atoms in Algorithm~\ref{algo:local_solution_mixture} are deferred to\comment{Section~\ref{Section:finite_mixture_composite} in} the Supplementary Material. Given the formulation of Algorithm~\ref{algo:local_solution_mixture},
we have the following result regarding its convergence to a local optimum.
\begin{thm}
\label{theorem:local_convergence_compo_dist_mixture} The Algorithm~\ref{algo:local_solution_mixture}
monotonically decreases the objective function~(\ref{eq:regu_optimal_compo})
of the regularized composite transportation distance for finite mixtures. 
\end{thm}
\subsection{Composite transportation barycenter for mixtures of exponential families}
In this section, we consider a problem of finding composite transportation barycenters for a collection of mixtures of exponential family. For $J \geq 1$, let $\left\{ P_{\bomega_{K_{j}}^{j},\bTheta_{K_{j}}^{j}}^{j}\right\} _{j=1}^{J}$
be a collection of $J$ mixtures of exponential families as described in~\eqref{eq:EF_Prob}, and let $\left\{ a_{j}\right\} _{j=1}^{J} \in \Delta^{J}$ be weights associated with these mixtures. The transportation barycenter of these probability measures is a mixture of exponential family with at most $L$ components, and is defined as an optimal solution of the following problem:
\begin{align}
\argmin{\bw_{L}, \boldsymbol{\Psi}_{L}} & \sum_{j=1}^{J}a_{j}\compoop\left(Q_{\bw_{L},\boldsymbol{\Psi}_{L}},P_{\bomega_{K_{j}}^{j},\bTheta_{K_{j}}^{j}}^{j}\right),\label{eq:MEF_Wass_barycenter}
\end{align}
where $\bw_{L} = \left\{ w_{l}\right\} _{l = 1}^{L} \in \simplex^{L}$ and
$\boldsymbol{\Psi}_{L}=\left\{ \psi_{l}\right\} _{l = 1}^{L}\in\Theta^{L}$ are unknown weights and parameters that we need to optimize. Recall that, to avoid notational clutter, we use $Q_{\bw_{L},\boldsymbol{\bPsi}_{L}}$ to denote a finite mixture with at most $L$ components. Since $P_{\bomega_{K_{j}}^{j},\bTheta_{K_{j}}^{j}}^{j}$ and $Q_{\bw_{L},\boldsymbol{\Psi}_{L}}$ are mixtures of exponential families, Eq. (\ref{eq:MEF_Wass_barycenter}) can be rewritten as
\begin{align*}
\argmin{\bw_{L},\boldsymbol{\Psi}_{L}} & \sum_{j=1}^{J}a_{j}\min_{\bpi^{j} \in \Pi\left(\bomega_{K_{j}}^{j},\bw_{L} \right)}\left\langle \bpi^{j},\bgamma^{j}\right\rangle,
\end{align*} where the cost matrices $\bgamma^{j} = (M_{uv}^{j})$ satisfy $M_{uv}^{j} = \KL(f(x|\theta_{u}^{j}),f(x|\psi_{v}))$ for $1 \leq j \leq J$, which has the closed form defined in Eq. (\ref{eq:cost_matrix})  since $f$ is from an exponential family of distributions. \comment{For the clarify of presentation, detail forms of $\bgamma^{j}$ are postponed to Section~\ref{Section:composite_transport_barycenter} in the Supplementary Material.}
\subsubsection{Regularized composite transportation barycenter} We incorporate regularizers in the composite transportation barycenter. In particular, we
write the objective function to be minimized as
\begin{align}
\argmin{\bw_{L},\boldsymbol{\Psi}_{L}} & \sum_{j=1}^{J}a_{j}\min_{\bpi^{j} \in \Pi\left(\bomega_{K_{j}}^{j},\bw_{L} \right)}\left\langle \bpi^{j},\bgamma^{j}\right\rangle - \lambda\entro\left(\bpi^{j}\right). \label{eq:regu_compo_bary}
\end{align}
We call this  objective function the \textit{regularized composite transportation barycenter}. Due to space constraints, we present the detailed algorithm for determining local solutions of this objective function\comment{ in Algorithm~\ref{algo:local_solution_barycenter}} in the Supplementary Material.

\section{Probabilistic clustering with multilevel structural data}
\label{sec:model}

Assume that we have $J$ groups of independent data, $X_{j,i}$, where $1 \leq j \leq J$ and $1 \leq i \leq n_{j}$; i.e., the data are presented in a two-level grouping structure. Our goal is to find simultaneously the local clustering for each data group and the global clustering across groups. 
\subsection{Multilevel composite transportation (MCT)} To facilitate the discussion, for each $1 \leq j \leq J$, we denote the empirical measure associated with group $j$ as
\begin{align*}  
P_{n_{j}}^{j} : = \frac{1}{n_{j}}\sum_{i=1}^{n_{j}} \delta_{X_{j,i}}. 
\end{align*} 
Additionally, we assume that the number of local and global clusters are bounded. In particular, we allow local group $j$ to have at most $K_{j}$ clusters, which can be represented as a mixture of exponential families $P_{\bomega_{j},\Theta_{K_{j}}^{j}}^{j}$, while we have at most $\gloclus$ global clusters among $J$ given groups. Here, each global cluster can be represented as a finite mixture distribution $Q_{\bw_{L}^{m},\bPsi_{L}^{m}}^{m}$ with at most $L$ clusters, where $\bw_{L}^{m} = (w_{1}^{m},\ldots,w_{L}^{m})$ and $\bPsi_{L}^{m} = (\psi_{1}^{m},\ldots,\psi_{L}^{m})$ are  global weights and atoms for $1 \leq m \leq \gloclus$, respectively.
\subsubsection{Local clustering and global clustering} With the local clustering, we perform composite transportation distance optimization for group $j$, which can be expressed as in~\eqref{eq:optimal_compo}. More precisely, this step can be viewed as finding optimal local weights $\bomega_{K_{j}}^{j}$ and local atoms $\Theta_{K_{j}}^{j}$ to minimize the composite transportation distance $\compoop(P_{n}^{j}, P_{\bomega_{K_{j}}^{j}, \bTheta_{K_{j}}^{j}})$ for all $1 \leq j \leq J$. Regarding the global clustering with $J$ given groups, we can treat the finite mixture probability distribution $P_{\bomega_{K_{j}}^{j}, \bTheta_{K_{j}}^{j}}$ of each group as observations in the space of distributions over probability distributions. Thus we achieve a clustering of these distributions by means of an optimization with the following composite transportation distance on the space of finite mixtures of finite mixtures:\useshortskip\begin{eqnarray}
\inf \limits_{\glomeas} \compoop\left(\locmeas, \glomeas\right), \nonumber
\end{eqnarray} where we denote $\locmeas : = \frac{1}{J}\sum_{j=1}^{J}\delta_{P_{\bomega_{K_{j}}^{j},\bTheta_{K_{j}}^{j}}^{j}}$ and $\glomeas := \sum_{m=1}^{\gloclus} b_{m} \delta_{Q_{\bw_{L}^{m},\bPsi_{L}^{m}}^{m}}$. 
\subsubsection{MCT formulation}
Since the finite mixture probability distributions  $P_{\bomega_{K_{j}}^{j}, \bTheta_{K_{j}}^{j}}$ in each group are unobserved, we determine them by minimizing the objective cost functions in the local clustering and global clustering simultaneously. In particular, we consider the following objective function:
\begin{align}
\inf \limits_{\substack{\bomega_{K_{j}}^{j}, \bTheta_{K_{j}}^{j},\glomeas}} \sum_{j=1}^{J} \compoop\left(P_{n_{j}}^{j}, P_{\bomega_{K_{j}}^{j},\bTheta_{K_{j}}^{j}}^{j}\right) + \zeta \compoop\left(\locmeas, \glomeas\right), \label{eq:multi_compo_optimal}
\end{align}
where $\zeta > 0$ serves as a penalization term between the global cluster and local cluster. We call this problem \textit{Multilevel Composite Transportation (MCT)}. 
\subsection{Regularized version of MCT} 
To obtain a favorable computation profile with MCT, we consider a regularized version of the composite transportation distances in both the local and global structures. To simplify the discussion, we denote $\bpi^{j} \in \Pi(\frac{1}{n_{j}}\boldsymbol{1}_{n_{j}}, \bomega_{K_{j}}^{j})$ as \textit{local transportation plans} between $P_{n_{j}}^{j}$ and $P_{\bomega_{K_{j}}^{j},\bTheta_{K_{j}}^{j}}^{j}$ for all $\leq j \leq J$.  Thus, the following formulation holds\begin{align}
\compoop\left(P_{n_{j}}^{j}, P_{\bomega_{K_{j}}^{j},\bTheta_{K_{j}}^{j}}^{j}\right) = \inf_{\bpi^{j} \in \Pi(\frac{1}{n_{j}}\boldsymbol{1}_{n_{j}}, \bomega_{K_{j}}^{j})} \left\langle \bpi^{j},\bgamma^{j} \right\rangle, \nonumber
\end{align} where $\bgamma^{j}$ is the cost matrix between $P_{n_{j}}^{j}$ and $P_{\bomega_{K_{j}}^{j},\bTheta_{K_{j}}^{j}}^{j}$ that is defined as\useshortskip\begin{align}
[M^{j}]_{uv} = - \log f(X_{j,u}|\theta_{v}^{j}), \nonumber
\end{align} for $1 \leq u \leq n_{j}$ and $1 \leq v \leq K_{j}$.
Therefore, we can consider the regularized version of composite transportation distance at each group $j$ as follows:
\begin{align}
\compoop\left(P_{n_{j}}^{j}, P_{\bomega_{K_{j}}^{j},\bTheta_{K_{j}}^{j}}^{j}\right) - \lambda_{l} \entro(\bpi^{j}),\label{eq:regu_local_structure}
\end{align} where $\lambda_{l} > 0$ is a penalization term for each group. Regarding the global structure, according to the definition of composite transportation distance for probability measure of measures, we have\begin{align}
\compoop\left(\locmeas, \glomeas\right) = \inf \limits_{\ba \in \Pi(\frac{1}{J}\boldsymbol{1}_{J}, \bb)} \sum_{j,m} a_{jm}\compoop(P_{\bomega_{K_{j}}^{j},\bTheta_{K_{j}}^{j}}^{j}, Q_{\bw_{L}^{m},\bPsi_{L}^{m}}^{m}), \nonumber
\end{align} where $\ba = (a_{jm})$ in the above infimum is a \textit{global transportation plan} between $\locmeas$ and $\glomeas$. Here, we can further rewrite $\compoop(P_{\bomega_{K_{j}}^{j},\bTheta_{K_{j}}^{j}}^{j}, Q_{\bw_{L}^{m},\Psi_{L}^{m}}^{m})$ as\begin{align}
\compoop(P_{\bomega_{K_{j}}^{j},\bTheta_{K_{j}}^{j}}^{j}, Q_{\bw_{L}^{m},\bPsi_{L}^{m}}^{m}) = \inf \limits_{\btau^{j,m} \in \Pi(\bomega_{K_{j}}^{j}, \bw_{L}^{m})} \left\langle \btau^{j,m},\bglo^{j,m} \right\rangle, \nonumber
\end{align} where the cost matrix $\bglo^{j,m}$ is defined as KL divergence between two exponential family atoms in Eq.~\eqref{eq:cost_matrix}:
\begin{align}\gamma_{k,l}^{j,m} & : = A\left(\theta_{k}^{j}\right) - A\left(\psi_{l}^{m} \right) \nonumber  - \left\langle \grad A\left(\psi_{l}^{m}\right),\left(\theta_{k}^{j} - \psi_{l}^{m} \right)\right\rangle. \nonumber
\end{align} To facilitate the discussion later, we denote $\btau^{j,m}$ the \textit{partial global transportation plan} between $P_{\bomega_{K_{j}}^{j}, \bTheta_{K_{j}}^{j}}$ and $Q_{\bw_{L}^{m},\Psi_{L}^{m}}^{m}$. Therefore, we can regularize the composite transportation distance with global structure as
\begin{align}
\compoop\left(\locmeas, \glomeas\right) - \lambda_{a}\entro(\ba) -\lambda_{g} \sum_{j=1}^{J}\sum_{m=1}^{\gloclus} \entro(\btau^{j,m}), \label{eq:regu_global_structure}
\end{align} where $\lambda_{a}$ corresponds to a penalization term for global structure while $\lambda_g$ represent a penalization term for a partial global transportation plan. Combining the results from Eqs.~\eqref{eq:regu_local_structure} and~\eqref{eq:regu_global_structure}, we obtain the overall objective function of MCT:
\begin{align}
& \inf \limits_{\bomega_{K_{j}}^{j}, \bTheta_{K_{j}}^{j}, \glomeas} \sum_{j=1}^{J} \compoop\left(P_{n_{j}}^{j}, P_{\bomega_{K_{j}}^{j},\bTheta_{K_{j}}^{j}}^{j}\right) + \zeta \compoop\left(\locmeas, \glomeas\right) - R\left(\bpi,\btau,\ba\right),  \label{eq:regu_multi_compo_optimal}
\end{align} where $R\left(\bpi,\btau,\ba\right) : = \lambda_{l} \sum_{j=1}^{J} \entro(\bpi^{j}) + \zeta [\lambda_{a} \entro(\ba) + \lambda_{g}\sum_{j=1}^{J}\sum_{m=1}^{\gloclus} \entro(\btau^{j,m})]$ is a combination of all regularized terms for the local and global clustering. We call this objective function \textit{regularized MCT}.
\subsection{Algorithm for regularized MCT}
\begin{algorithm}
\begin{algorithmic}   
\REQUIRE Data $D=\left\{ X_{j,i}\right\} _{j=1,i=1}^{J,n_{j}}$; the number of local clusters $K_{j}$ and global clusters $\gloclus$; the number of components in each global cluster $L$; the penalization term $\zeta$; the regularized hyper-parameters $\lambda_{l}, \lambda_{g}$.
\ENSURE local and global parameters $\bomega_{K_{j}}^{j}, \bTheta_{K_{j}}^{j}, \bw_{L}^{m}, \bPsi_{L}^{m}$ for $1 \leq j \leq J$ and $1 \leq m \leq \gloclus$.    
\STATE Initialize these local  and global parameters.   
\WHILE{not converged}
\STATE 1. Update local parameters:.    
\FOR{$j=1$ \TO $J$}    
\STATE Update $\bomega_{K_{j}}^{j}$ , $\bTheta_{K_{j}}^{j}$ as optimal solutions of~\eqref{eq:local_clustering_updates}. 
\ENDFOR    
\STATE 2. Update global parameters:   
\FOR{$m=1$ \TO $\gloclus$}      
\STATE Update $\bw_{L}^{m}, \bPsi_{L}^{m}$ as optimal solutions of~\eqref{eq:global_clustering_updates}.   
\ENDFOR 
\ENDWHILE 
\end{algorithmic}
\caption{\label{algorithm:multilevel_composite_transportation_updates}
Probabilistic Multilevel Clustering}
\end{algorithm}
We now describe our detailed strategy for obtaining a locally optimal solution of \textit{regularized MCT}. In particular, our algorithm consists of two key steps: local clustering updates and global clustering updates. For simplicity of presentation, we assume that at step $t$ of our algorithm, we have the following updated values of our parameters: $\bomega_{K_{j}}^{j}, \bTheta_{K_{j}}^{j}, \bw_{L}^{m}, \bPsi_{L}^{m}$, for $1 \leq j \leq J$ and $1 \leq m \leq \gloclus$.
\paragraph{Local clustering updates}
To obtain updates for local weights $\bomega_{K_{j}}^{j}$ and local atoms $\bTheta_{K_{j}}^{j}$, we solve the following combined regularized composite transportation barycenter problem:
\begin{align}
& \hspace{ - 5 em} \inf \limits_{\bomega_{K_{j}}^{j}, \bTheta_{K_{j}}^{j}} \compoop\left(P_{n_{j}}^{j}, P_{\bomega_{K_{j}}^{j},\bTheta_{K_{j}}^{j}}^{j}\right) - \lambda_{l} \entro(\bpi^{j})  \label{eq:local_clustering_updates}\\
& + \sum_{m = 1}^{\gloclus} a_{jm}\compoop(P_{\bomega_{K_{j}}^{j},\bTheta_{K_{j}}^{j}}^{j}, Q_{\bw_{L}^{m},\Psi_{L}^{m}}^{m}) - \zeta \lambda_{g}\sum_{m=1}^{\gloclus} \entro(\btau^{j,m}), \nonumber
\end{align} where $\bpi^{j}$ is the local transportation plan between $P_{n_{j}}^{j}$ and $P_{\bomega_{K_{j}}^{j},\bTheta_{K_{j}}^{j}}^{j}$ at step $t$ while $\ba$ and $\btau^{j,m}$ are respectively the global transportation plan and partial global transportation plans at this step. The idea of obtaining local solution of the above objective function is identical to that of~\eqref{eq:regu_compo_bary}; therefore, we defer the detailed presentation of this algorithm\comment{in Section~\ref{Section:multilevel_clustering_updates}} to the Supplementary Material.
\paragraph{Global clustering updates} In order to update the global weights $\bw_{L}^{m}$ and global atom parameters $\bPsi_{L}^{m}$, we consider the following optimization problem:\begin{align}
\inf \limits_{\glomeas} \compoop\left(\locmeas, \glomeas\right) - \zeta [\lambda_{a}  \entro(\ba) + \lambda_{g}\sum_{j=1}^{J}\sum_{m=1}^{\gloclus} \entro(\btau^{j,m})]. \label{eq:global_clustering_updates}
\end{align} The algorithm for obtaining the local solutions of this objective is based on bacycenter computation algorithms in \cite{Cuturi-2014} for updating barycenter weights $\bw_{L}^{m}$ and  the partial global transportation plan $\btau^{j,m} $.  The natural parameters of global atoms of the barycenters are weighted averages of local atoms from all group $j$: \begin{align}
\psi_{v} = \frac{\sum_{j = 1}^{J} \sum_{u = 1}^{L} \pi_{uv}^{j} \theta_{u}^{j}}{\sum_{j = 1}^{J} \sum_{u = 1}^{L} \pi_{uv}^{j}} \nonumber.
\end{align} The detailed derivation of this algorithm is deferred to\comment{Section~\ref{Section:multilevel_clustering_updates} in} the Supplementary Material. In summary, the main steps of updating the local and global clustering updates are summarized in Algorithm~\ref{algorithm:multilevel_composite_transportation_updates}. We have the following result guaranteeing the local convergence of this algorithm.
\begin{thm}
\label{theorem:local_convergence_multilevel_composite} Algorithm~\ref{algorithm:multilevel_composite_transportation_updates}
monotonically decreases the objective function of regularized MCT ~\eqref{eq:regu_multi_compo_optimal}
until local convergence. 
\end{thm}

\section{Experimental studies} 
\label{sec:experiments}

\begin{table*}
\begin{centering}
\setlength\tabcolsep{1pt}\subfloat[\label{tab:rsynthetic_dataset_statistics}Statistics of synthetic datasets]{\begin{centering}
\begin{tabular}{|c|c|c|c|c|}
\hline 
Datasets & \multicolumn{1}{c||}{\textcolor{black}{\#groups(J)}} & \multicolumn{1}{c||}{\#dim} & \multicolumn{1}{c||}{\#points($n_{j})$} & \textcolor{black}{\#clusters($\gloclus$)}\tabularnewline
\hline 
\hline 
Continuous data & $100$ & $2$ & $500$ & $6$\tabularnewline
\hline 
Discrete data & $500$ & $25$ & $100$ & $5$\tabularnewline
\hline 
\end{tabular}
\par\end{centering}

}\quad{}\subfloat[\label{tab:realworld-datasets_stats}Statistics of real-world datasets]{\begin{centering}
{\small{}}%
\begin{tabular}{|c|c|c|c|}
\hline 
{\small{}Datasets} & \textcolor{black}{\small{}\#groups(J)} & {\small{}\#dim} & \textcolor{black}{\small{}\#clusters($\gloclus$)}\tabularnewline
\hline 
\hline 
{\small{}LabelMe } & {\small{}$1,800$} & {\small{}$30$} & {\small{}$8$}\tabularnewline
\hline 
{\small{}NUS-WIDE} & {\small{}$1,040$} & {\small{}$238$} & {\small{}$13$}\tabularnewline
\hline 
\end{tabular}
\par\end{centering}{\small \par}

}
\par\end{centering}

\caption{Summarization of synthetic and realworld datasets }
\end{table*}We first evaluate the model via simulation studies, then demonstrate its applications on text and image modeling using two real-world datasets.
\subsection{Simulated data}

We evaluate the
effectiveness of our proposed clustering algorithm by
considering two types (discrete and continuous) of synthetic data generated from multilevel processes as follows.

\paragraph{Continuous data}

\begin{figure}
\begin{centering}
\includegraphics[width=1\columnwidth]{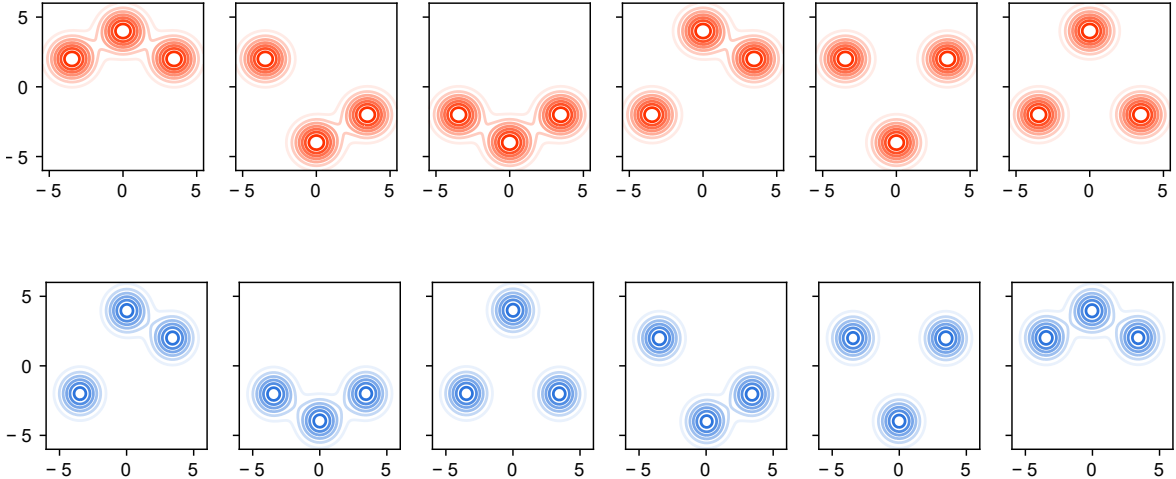}
\par\end{centering}

\caption{\label{fig:syn_GMM}Synthetic multilevel Gaussian data (orange at top)
and inferred clusters (blue at bottom).}

\end{figure}

We start with six clusters of data, each of which is a mixture of three Gaussian components.
Figure \ref{fig:syn_GMM} depicts the ground truth of the six mixtures we generate the data from. We uniformly generated $100$ groups of data, each group belonging to one of the six aforementioned clusters. Once the cluster index of a data group was defined, we generated $500$ data points from the corresponding mixture of Gaussians.

\paragraph{Discrete data}

\begin{figure}
\begin{centering}
\includegraphics[width=1\columnwidth]{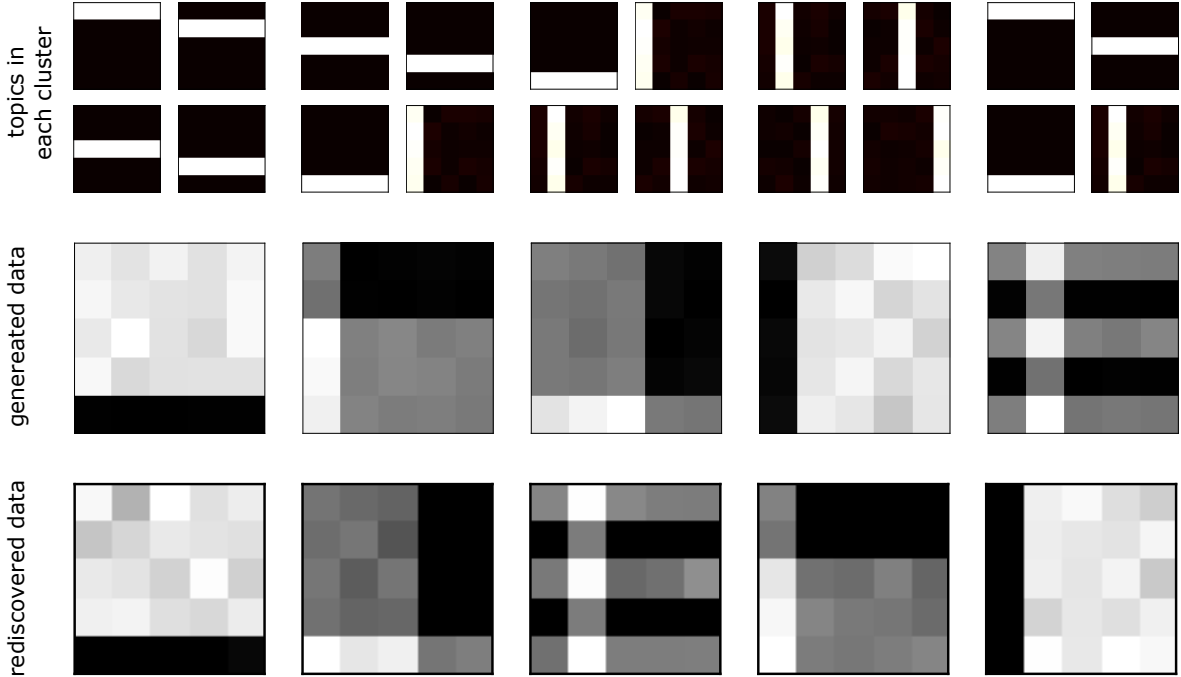}
\par\end{centering}

\caption{\label{fig:syn_bartopic}Synthetic multilevel bar topic data (two
top rows) and rediscovered output (bottom row)}
\end{figure}

Data was generated from five clusters of $25$-dimensional bar topics, each of which is a mixture of four bar topics out of total ten topics as shown in Figure \ref{fig:syn_bartopic} (second row). Each cluster  shares two topics with any other cluster. We then generated $500$ groups of data, each group belonging to one of the five aforementioned clusters. Once the cluster index of a data group is defined, we generate $100$ data points from the mixture of bar topics of that cluster.

\paragraph{Clustering results }
We ran the proposed method with \emph{synthetic continuous data} using the following local and global penalization hyper-parameters: $\lambda_{l}$ and $\lambda_{g}$ are set equal to $1.3$ and $10$, respectively. We model each atom in the (local and global) mixture models as an isotropic multivariate Gaussian. As shown in the bottom row of Figure \ref{fig:syn_GMM}, the model is able to rediscover the clustering structure in the generated dataset. Comparing with the top row of the figure, there is permutation in the order of discovered mixture models due to the label switching.
Similarly, we use Categorical distribution to model each atom in the mixture models of the proposed model. Each observation $X_{ji}$ now is a one-hot vector. We simulated from ten topics including five horizontal and five vertical bars. The top row of Figure \ref{fig:syn_bartopic} depicts a collection of four bar topics that data of a cluster may be generated from; i.e., the first cluster contains data simulated from a mixture of four horizontal bar topics. In the middle row, we depict the histogram plot of all data generated from each cluster while the bottom row shows the plot of clusters discovered by our proposed model. There is only a slight difference in the plot between ground truth and the inferred mixture of bar topics\footnote{We also measured the NMI (Normalized Mutual Information) between the ground truth labels and learned groups and obtained around $0.98$. }. These results demonstrate the effectiveness and flexibility of our algorithms in learning both continuous and discrete data.
\subsection{Real-world data}
We now demonstrate our proposed model on two real-world datasets:
the LabelMe dataset \cite{russell2008labelme,Oliva-2001} with continuous observations and the NUS-WIDE \cite{chua2009nus} with discrete observations. Statistics for these datasets are presented in Table \ref{tab:realworld-datasets_stats}.

\paragraph{LabelMe dataset}
This consists of $2,688$ annotated images which are classified into eight
scene categories including \emph{tall buildings, inside city, street,
highway, coast, open country, mountain, and forest} \cite{russell2008labelme}.
Each image contains multiple annotated regions. Each region, which
is annotated by users, represents an object in the image\footnote{Sample images and annotated regions can be found at http://people.csail.mit.edu/torralba/code/spatialenvelope/}. We remove the images containing less than four annotated regions and obtained totally $1,800$ images. We then extract GIST features~\cite{lowe1999object}, a visual descriptor to represent perceptual dimensions and oriented spatial structures of a scene, for each region in an image. We use PCA to reduce the number of dimensions to 30.

\begin{table}
\begin{centering}
\begin{tabular}{|c|c|c|c|}
\hline 
Methods & \textcolor{black}{NMI} & ARI & \textcolor{black}{AMI}\tabularnewline
\hline 
\hline 
\textcolor{black}{K-means} & 0.37 & 0.282 & 0.365\tabularnewline
\hline 
SVB-MC2 & 0.315 & 0.206 & 0.273\tabularnewline
\hline 
\textcolor{black}{W-means} & 0.423 & 0.35 & 0.416\tabularnewline
\hline 
\textbf{MCT} & \textbf{0.485} & \textbf{0.412} & \textbf{0.477}\tabularnewline
\hline 
\end{tabular}
\par\end{centering}

\caption{\label{tab:labelme_clustering_metrics}Clustering performance on LabelMe
(continuous) dataset.}
\end{table}
\begin{table}
\begin{centering}
\begin{tabular}{|c|c|c|c|}
\hline 
Methods & \textcolor{black}{NMI} & ARI & \textcolor{black}{AMI}\tabularnewline
\hline 
\hline 
\textcolor{black}{K-means} & 0.35  & 0.093 &  0.22\tabularnewline
\hline 
SVB-MC2 & 0.295 & 0.139 & 0.249\tabularnewline
\hline 
\textcolor{black}{W-means} & 0.356 & 0.089 & 0.203\tabularnewline
\hline 
\textbf{MCT} & \textbf{0.423} & \textbf{0.255} & \textbf{0.39}\tabularnewline
\hline 
\end{tabular}
\par\end{centering}

\caption{\label{tab:nuswide_clustering_metrics}Clustering performance on NUS-WIDE
(discrete) dataset.}
\end{table}

\paragraph{NUS-WIDE dataset}
We used a subset of the original NUS-WIDE dataset \cite{chua2009nus}
which contains images of 13 kinds of animals\footnote{including squirrel, cow, cat, zebra, tiger, lion, elephant, whales, rabbit, snake, antlers, hawk and wolf} comprising $2,054$ images in training subset. Each image is annotated with several tags out of $1,000$ tags. We filtered out images with less than three tags and obtained $1,040$ images with the remaining number tags of $238$. After preprocessing, we have a dataset with $1,040$ groups; each data point in a group is a one-hot vector of 238 dimensions
representing a tag word annotated for that group (image).
\begin{figure}
\begin{centering}
\includegraphics[width=1\columnwidth]{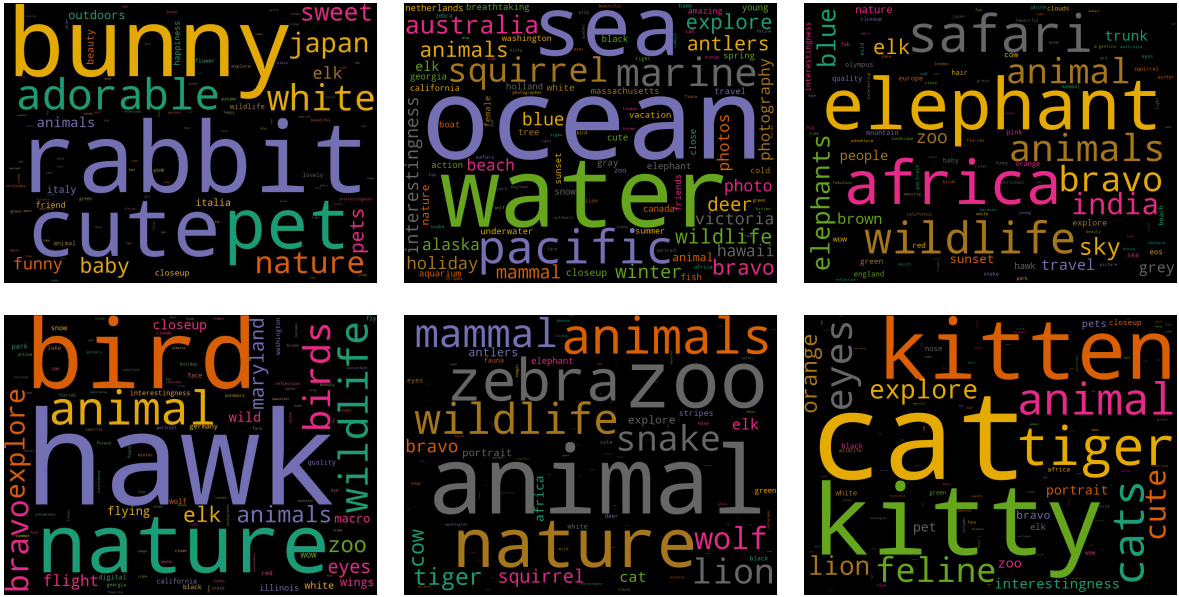}
\par\end{centering}

\caption{\label{fig:nuswide_wordcloud}Tag cloud of six clusters
discovered by the proposed model with the NUS-WIDE dataset.}
\end{figure}

\paragraph{Baseline methods }

We \emph{quantitatively} compare our proposed method to
baseline approaches discussed in \cite{ho17multilevel}, including
\emph{K-means}, \emph{W-means}, and \emph{SVB-MC2 without
context }\cite{huynh2016scalable}. We use three popular metrics: NMI (Normalized Mutual Information) \cite[16.3]{schutze2008introduction}, ARI (Adjusted
Rand Index) \cite{hubert1985comparing}, and AMI (Adjusted Mutual
Information) \cite{vinh2010information} to evaluate the clustering
performance.

\paragraph{Experimental results}

We conducted experiments on the LabelMe dataset with the number of local atoms set equal to $K=5$, the number of global atoms set to $L=15$, and the number of clusters set to $\gloclus=8$. We chose the hyper-parameters for penalized terms to be $\lambda_{l}=3$ and $\lambda_{g}=3$. As shown in Table \ref{tab:labelme_clustering_metrics}, our proposed method is superior to the baseline methods in terms of clustering performance. 

We also compared clustering performance using the discrete real-world dataset NUS-WIDE. We chose $K=2$, $L=4$, and $\gloclus=13$ with the hyper-parameters $\lambda_{l}=1$ and $\lambda_{g}=1.6$. Results are presented in Table \ref{tab:nuswide_clustering_metrics}. Since baseline methods are applicable only to continuous data, we have normalized the discrete data for each image and then applied the baseline methods to cluster the dataset. The results show that the clustering performance of K-means and W-means is inferior to that of our proposed model which directly models discrete data. Moreover, the Adjusted Rand Index (ARI) and Adjusted Mutual Information (AMI) of these model show that their clustering outcomes are not robust. 

To illustrate the \textit{qualitative} results of the proposed model, we selectively choose six out of thirteen clusters discovered and computes the proportion of tags presented in each cluster. Figure \ref{fig:nuswide_wordcloud} depicts tag-clouds of these clusters. Each tag-cloud consistently manifests the cluster content. For example, the top-left tag-cloud denote the cluster of rabbits, which is one of the ground-truth subsets of images.

\section{Discussion}
\label{sec:conclusion}

We have proposed a probabilistic model that uses a novel composite transportation distance to cluster  data with potentially complexed hierarchical multilevel structures. The proposed model is able to handle both discrete and continuous observations. Experiments on simulated and real-world data have shown that our approach outperforms competing methods that also target multilevel clustering tasks. Our developed model is based on the exponential family assumption with data distribution and thereby applies naturally to other data types; e.g., a mixture of Poisson distributions \cite{jayasekare2016modeling}. Finally, there are several possible directions for extensions from our work. First, it is of interest to extend our approach to richer settings of hierarchical data similar to those considered in MC$^2$ \cite{Vu-2014}; e.g., when group-level context is available in the data. Second, our method requires knowledge of the upper bounds with the numbers of clusters both in local and global clustering. It is of practical importance to develop methods that are able to estimate these cardinalities efficiently. 

\bibliographystyle{abbrv}
\bibliography{Nhat,optimaltransport}

\clearpage{}
\begin{center}
\LARGE{\bf{Appendix}}
\end{center}
\appendix

\section{Finite mixtures with regularized composite transportation distance}
\label{Section:finite_mixture_composite}
In this section, we provide detailed analyses for obtaining updates with weights and atoms in Algorithm~\ref{algo:local_solution_mixture} to find the local solution of the objective function in Eq.~\eqref{eq:regu_optimal_compo}, which optimizes finite mixtures with regularized composite transportation distance. To ease the presentation, we would like to remind this objective function, which is defined as follows
\begin{align*}
\min \limits_{\bomega_{K}, \bTheta_{K}} \inf_{\bpi\in\Pi\left(\frac{1}{n}\boldsymbol{1}_{n},\bomega_{K} \right)} \left\langle \bpi,\bgamma\right\rangle - \lambda\entro\left(\bpi\right)
\end{align*}
where $\lambda>0$ is a penalization term and $\entro\left(\bpi\right) = - \sum_{i,j} \pi_{ij}\log \pi_{ij}$ is an entropy of $\bpi \in \Pi\left(\boldsymbol{1}_{n}/n,\bomega_{K} \right)$. Here, $P_{n} = \frac{1}{n} \sum_{i=1}^{n} \delta_{X_{i}}$ an empirical measure with respect to samples $X_{1},\ldots,X_{n}$. Furthermore, $\bgamma = (M_{ij})$ is a cost matrix such that $M_{ij} = - \log f(X_{i}| \theta_{j})$ for $1 \leq i \leq n, 1 \leq j \leq K$ while $\Pi\left(\cdot,\cdot\right)$ is the set of transportation plans between $\boldsymbol{1}_{n}/n$ and $\bomega_{K}$.
\subsection{Update weights}
\label{subsection:weights_update_mixtures}
Our strategy for updating weights $\bomega_{K}$ in the above objective function relies on solving the
following relaxation of that optimization problem\begin{align}
\inf\limits _{\bpi \in \mathcal{S}_{n}}\left\langle \bpi,\bgamma\right\rangle -\lambda\entro\left(\bpi\right)\label{eqn:weight_update_mixture}
\end{align}where $\mathcal{S}_{n} = \left\{\bpi: \ \sum_{j=1}^{K} \pi_{ij} = 1/n \right\}$. Invoking the Lagrangian multiplier for the constraint $\bpi\boldsymbol{1}_{K}=\frac{1}{n}\boldsymbol{1}_{n}$,
the above objective function is equivalent to minimize the following function\begin{align*}
\mathcal{F} & =\sum_{i = 1}^{n} \sum_{j = 1}^{K} \pi_{ij}M_{ij} + \lambda\sum_{i = 1}^{n}\sum_{j = 1}^{K}\pi_{ij}\left(\log \pi_{ij}-1\right) + \sum_{i = 1}^{n} \kappa_{i}\left(\sum_{j = 1}^{K} \pi_{ij}-\frac{1}{n}\right).
\end{align*} By taking the derivative of $\mathcal{F}$ with respect to $\pi_{ij}$ and setting it to zero, the following equation holds\begin{align*}
\partialdev{\mathcal{F}}{\pi_{ij}} & = M_{ij}+ \lambda \log \pi_{ij} + \kappa_{i} = 0.
\end{align*} The above equation leads to\begin{align*}
\pi_{ij} & =\exp\left(\frac{-M_{ij}-\kappa_{i}}{\lambda}\right)=\left(f(X_{i}|\theta_{j})\right)^{1/\lambda}\exp\left(\frac{-\kappa_{i}}{\lambda}\right).
\end{align*} Invoking the condition $\sum_{k}\pi_{ik}=\frac{1}{n}$, we have
\begin{align*}
\exp\left(\frac{-\kappa_{i}}{\lambda}\right)\left(\sum_{j = 1}^{K} \left(f(X_{i}|\theta_{j})\right)^{1/\lambda}\right) & =\frac{1}{n},
\end{align*}
which suggests that\begin{align*}
\exp\left(\frac{-\kappa_{i}}{\lambda}\right) & =\frac{1}{n}\frac{1}{\sum_{j = 1}^{K} \left(f(X_{i}|\theta_{j})\right)^{1/\lambda}}.
\end{align*} Governed by the previous equations, we find that\begin{align}
\pi_{ij}= & \frac{1}{n}\frac{\left(f(X_{i}|\theta_{j})\right)^{1/\lambda}}{\sum_{j = 1}^{K}\left(f(X_{i}|\theta_{j})\right)^{1/\lambda}}.\label{eqn:transport_plan_update}
\end{align} Therefore, we can update the weight $\omega_{j}$ as\begin{align}
\omega_{j}=\sum_{i=1}^{n}\pi_{ij} \nonumber
\end{align} for any $1\leq j \leq K$.
\subsection{Update atoms}
\label{subsection:atoms_update_mixtures}
Given the updates for weight $\bomega_{K}$ and the formulation of  cost matrix $\bgamma$, to obtain the update
for atoms $\theta_{j}$ as $1\leq j \leq K$, we optimize the following objective function 
\begin{align}
\min_{\bTheta_{K}}-\sum_{i = 1}^{n} \sum_{j = 1}^{K} & \pi_{ij}\log f(X_{i}|\theta_{j}). \label{eqn:atom_update_mixture}
\end{align}
Since $f(x|\theta)$ is an exponential family distribution with natural
parameter $\theta$, we can represent it as 
\begin{align*}
f(x|\theta) & = h(x)\exp\left(\left\langle T\left(x\right),\theta\right\rangle -A\left(\theta\right)\right),
\end{align*}
where $A\left(\theta\right)$ is the log-partition function which
is convex. Plugging this formulation of $f(x|\theta)$ into the objective function~\eqref{eqn:atom_update_mixture} and taking the derivative with respect to $\theta_{j}$, we obtain the following equation
\begin{align}
\sum_{i = 1}^{n} \pi_{ij} T(X_{i}) - \omega_{j}\grad A(\theta_{j}) = 0. \label{eqn:atom_update}
\end{align}
Therefore, we can update atoms $\theta_{j}$ as the solution of the above equation for $1 \leq j \leq K$,
\subsection{Proof for local convergence of Algorithm~\ref{algo:local_solution_mixture}}
Given the formulation of Algorithm~\ref{algo:local_solution_mixture},
we would like to demonstrate its convergence to local solution of objective function~\eqref{eq:regu_optimal_compo} in Theorem~\ref{theorem:local_convergence_compo_dist_mixture}.

Our proof of the theorem is straight-forward from the updates of weights
and atoms via Lagrangian multipliers. In particular, we denote $\bomega_{K}^{(t)}$,
$\bTheta_{K}^{(t)}$, and $\bpi^{(t)}$ as the update of weights,
atoms, and transportation plan in step $t$ of Algorithm~\ref{algo:local_solution_mixture}
for $t\geq0$. Additionally, let $\bgamma^{(t)}$ be the cost matrix
at step $t$, i.e., $\bgamma_{ij}^{(t)}=-\log f(X_{i}|\theta_{j}^{(t)})$
for all $i,j$. Furthermore, we denote 
\begin{align*}
g(\bomega_{K},\bTheta_{K}):=\inf_{\bpi\in\Pi\left(\frac{1}{n}\boldsymbol{1}_{n},\bomega_{K}\right)}\left\langle \bpi,\bgamma\right\rangle -\lambda\entro\left(\bpi\right).
\end{align*}
Then, for any $t\geq0$, it is clear that 
\begin{align*}
g(\bomega_{K}^{(t)},\bTheta_{K}^{(t)}) & =\inf\limits _{\bpi\in\Pi\left(\frac{1}{n}\boldsymbol{1}_{n},\bomega_{K}^{(t)}\right)}\left\langle \bpi,\bgamma^{(t)}\right\rangle -\lambda\entro\left(\bpi\right)\\
 & \geq\inf\limits _{\bpi\in\mathcal{S}_{n}}\left\langle \bpi,\bgamma^{(t)}\right\rangle -\lambda\entro\left(\bpi\right)\\
 & \geq\left\langle \bpi^{(t+1)},\bgamma^{(t)}\right\rangle -\lambda\entro\left(\bpi^{(t+1)}\right)
\end{align*}
where $\mathcal{S}_{n}=\left\{ \bpi:\sum_{j=1}^{K}\pi_{ij}=1/n\ \forall1\leq i\leq n\right\} $.
Here, the first inequality is due to the fact that $\Pi\left(\frac{1}{n}\boldsymbol{1}_{n},\bomega_{K}^{(t)}\right)\subset\mathcal{S}_{n}$
while the second inequality is due to~\eqref{eqn:transport_plan_update} in subsection~\ref{subsection:weights_update_mixtures}.
According to the update of atoms in~\eqref{eqn:atom_update} in subsection~\ref{subsection:atoms_update_mixtures},
we have that 
\begin{align*}
\left\langle \bpi^{(t+1)},\bgamma^{(t)}\right\rangle -\lambda\entro\left(\bpi^{(t+1)}\right) & \geq \left\langle \bpi^{(t+1)},\bgamma^{(t+1)}\right\rangle -\lambda\entro\left(\bpi^{(t+1)}\right)\\
 & \geq\inf\limits _{\bpi\in\Pi\left(\frac{1}{n}\boldsymbol{1}_{n},\bomega_{K}^{(t+1)}\right)}\left\langle \bpi,\bgamma^{(t+1)}\right\rangle -\lambda\entro\left(\bpi\right)\\
 & =g(\bomega_{K}^{(t+1)},\bTheta_{K}^{(t+1)}).
\end{align*}
Governed by the above results, for any $t\geq0$, the following holds
\begin{align*}
g(\bomega_{K}^{(t)},\bTheta_{K}^{(t)})\geq g(\bomega_{K}^{(t+1)},\bTheta_{K}^{(t+1)}).
\end{align*}
As a consequence, we achieve the conclusion of Theorem~\ref{theorem:local_convergence_compo_dist_mixture}. 
\section{Regularized composite transportation barycenter}
\label{Section:composite_transport_barycenter}
In this section, we provide a detailed algorithm for achieving local solution to regularized composite transportation barycenter in objective function in Eq.~\eqref{eq:regu_compo_bary}. To facilitate the discussion, we will remind the formulation of that objective function. In particular, the objective function with regularized composite transportation distance has the following formulation\begin{align*}
\min \limits_{\bw_{L},\boldsymbol{\Psi}_{L}} & \sum_{j=1}^{J}a_{j}\min_{\bpi^{j} \in \Pi\left(\bomega_{K_{j}}^{j},\bw_{L} \right)}\left\langle \bpi^{j},\bgamma^{j}\right\rangle - \lambda\entro\left(\bpi^{j}\right)
\end{align*}
where $\bgamma^{j}$ is the corresponding KL cost matrix between finite mixture probability distribution $P_{\bomega_{K_{j}}^{j},\bTheta_{K_{j}}^{j}}^{j}$ and $Q_{\bw_{L},\boldsymbol{\Psi}_{L}}$ for $1 \leq j \leq J$. Here, $\left\{ a_{j}\right\} _{j=1}^{J} \in \Delta^{J}$ are
given weights associated with the finite mixture probability distributions $\left\{P_{\bomega_{K_{j}}^{j},\bTheta_{K_{j}}^{j}}^{j}\right\}_{j=1}^{J}$. 

As $f(x|\theta)$ is an exponential family, the cost matrix $\bgamma^{j} = (M_{uv}^{j})$ has the following formulation
\begin{align*}
M_{uv}^{j} & = \KL( f(x|\psi_{v}),f(x|\theta_{u}^{j})) \nonumber \\
& = A(\theta_{u}^{j}) - A(\psi_{v}) - \left\langle \grad A(\psi_{v}), (\theta_{u}^{j} - \psi_{v}) \right\rangle
\end{align*}
for all $1 \leq u,v \leq K$.
\subsection{Update weights and atoms}
Our procedure for updating weights $\bw_{K}$ for the objective function of regularized composite transportation distance will be similar to Algorithm 1 in~\cite{Cuturi-2014}. Therefore, we will only focus on the updates with atoms $\boldsymbol{\Psi}_{L} = (\psi_{1}, \ldots, \psi_{L})$. 

Given the updates of weights $\bw_{K}$, we compute the optimal transportation plan $\bpi^{j} = (\pi_{uv}^{j})$ between $\bomega_{K_{j}}^{j}$ and $\bw_{L}$ using  Algorithm 3 in~\cite{Cuturi-2014}. Then, to obtain the updates for $\boldsymbol{\Psi}_{L}$, we consider the following optimization problem\begin{align*}
\min \limits_{\boldsymbol{\Psi}_{L}} \sum_{j = 1}^{J} a_{j} \sum_{u = 1}^{K_{j}} \sum_{v = 1}^{K_{j}} \pi_{uv}^{j}M_{uv}^{j}.
\end{align*} 
By taking the derivative of the above objective function with respect to $\psi_{v}$ and setting it to 0, we achieve the following equation\begin{align}
\sum_{j = 1}^{J} \sum_{u = 1}^{K_{j}} \pi_{uv}^{j} \left\langle \grad^{2} A(\psi_{v}), \theta_{u}^{j} - \psi_{v} \right\rangle = 0. \nonumber 
\end{align} One possible solution to the above equation is $\sum_{j = 1}^{J} \sum_{u = 1}^{K_{j}} \pi_{uv}^{j} (\theta_{u}^{j} - \psi_{v}) = 0$. This previous equation suggests that
\begin{align}
\psi_{v} = \frac{\sum_{j = 1}^{J} \sum_{u = 1}^{K_{j}} \pi_{uv}^{j} \theta_{u}^{j}}{\sum_{j = 1}^{J} \sum_{u = 1}^{K_{j}} \pi_{uv}^{j}} \label{eq:atom_update_barycenter}
\end{align}
for all $1 \leq v \leq L$. Equipped with these updates for weights $\bw_{L}$ and atoms $\boldsymbol{\Psi}_{L}$, we summarize the detail of an algorithm for determining the local solution of regularized composite transportation barycenter in Eq. ~\eqref{eq:regu_compo_bary} in Algorithm~\ref{algo:local_solution_barycenter}. 
\begin{algorithm}
\begin{algorithmic} 
\REQUIRE
Finite mixture probability distributions $\left\{ P_{\bomega_{K_{j}}^{j},\bTheta_{K_{j}}^{j}}^{j}\right\} _{j=1}^{J}$, given weights $\{a_{j}\}_{j=1}^{J}$, and the regularized hyper-parameter $\lambda>0$. 
\ENSURE Optimal
weights $\bw_{L}$ and atoms $\boldsymbol{\Psi}_{L}$.
\STATE Initialize weights $\left\{ w_{j}\right\} _{j=1}^{L}$
and atoms $\left\{ \psi_{j}\right\} _{j=1}^{L}$. \WHILE{not converged}
\STATE 1. Update weights $\bw_{L}$ as Algorithm 1 in~\cite{Cuturi-2014}.
\STATE 2. Compute transportation plans $\bpi^j$ for $1\le j\le J$ using  Algorithm 3 in~\cite{Cuturi-2014}.
\STATE 3. Update atoms $\boldsymbol{\Psi}_{L}$ as in Eq.~\eqref{eq:atom_update_barycenter}.
\ENDWHILE \end{algorithmic} \caption{\label{algo:local_solution_barycenter} Regularized composite transportation barycenter}
\end{algorithm}
\subsection{Local convergence of Algorithm~\ref{algo:local_solution_barycenter}}
Given the formulation of Algorithm~\ref{algo:local_solution_barycenter}, the following theorem demonstrates that this algorithm determines the local solution of objective function~\eqref{eq:regu_compo_bary}
\begin{thm}
\label{theorem:local_convergence_compo_barycenter} The Algorithm~\ref{algo:local_solution_barycenter}
monotonically decreases the objective function~\eqref{eq:regu_compo_bary}
of regularized composite transportation barycenter
until local convergence. 
\end{thm}
The proof of Theorem~\ref{theorem:local_convergence_compo_barycenter} is a direct consequence of the updates with weights and atoms in the above subsection and can be argued in the similar fashion as that of Theorem~\ref{theorem:local_convergence_compo_dist_mixture}; therefore, it is omitted.
\section{Multilevel clustering with composite transportation distance}
\label{Section:multilevel_clustering_updates}
In this section, we provide detailed argument for the algorithm development to determine the local solutions of regularized multilevel composite transportation (MCT). To ease the presentation later, we would like to remind the objective function of this problem as well as all its important relevant notations. We start with the objective function in Eq.~\eqref{eq:regu_multi_compo_optimal} as follows\begin{align*}
& \inf \limits_{\bomega_{K_{j}}^{j}, \bTheta_{K_{j}}^{j}, \glomeas} \sum_{j=1}^{J} \compoop\left(P_{n_{j}}^{j}, P_{\bomega_{K_{j}}^{j},\bTheta_{K_{j}}^{j}}^{j}\right) + \zeta \compoop\left(\locmeas, \glomeas\right) - R\left(\bpi,\btau,\ba\right), 
\end{align*} 
where $R\left(\bpi,\btau,\ba\right)  = \lambda_{l} \sum_{j=1}^{J} \entro(\bpi^{j}) + \zeta \lambda_{a}[ \entro(\ba) + \lambda_{g}\sum_{j=1}^{J}\sum_{m=1}^{\gloclus} \entro(\btau^{j,m})]$ is a combination of all regularized terms for the local and global clustering. Here, for the simplicity of our argument, we choose $\zeta = 1$ to derive our learning updates. In the above objective function, $\locmeas  = \frac{1}{J}\sum_{j=1}^{J}\delta_{P_{\bomega_{K_{j}}^{j},\bTheta_{K_{j}}^{j}}^{j}}$ and $\glomeas = \sum_{m=1}^{\gloclus} b_{m} \delta_{Q_{\bw_{L}^{m},\bPsi_{L}^{m}}^{m}}$. We summarize below the notations for our algorithm development.
\subparagraph{Variables of local clustering structures }
\begin{itemize}
\item Local transportation plans for group $j$: $\bpi^{j}=\left\{ \pi_{uv}^{j}\right\} _{u,v} \in \Pi(\frac{1}{n_{j}}\boldsymbol{1}_{n_{j}}, \bomega_{K_{j}}^{j})$
s.t. $\sum_{v}\pi_{uv}^{j}=\frac{1}{n_{j}}$, and $\sum_{u}\pi_{uv}^{j}=\omega_{v}^{j}$,
\item Local atoms for local group $\boldsymbol{\Theta}_{K_{j}}^{j} = \left\{ \theta_{k}^{j}\right\} _{k = 1}^{K_{j}}$
and their local mixing weights $\bomega^{j}=\left\{ \omega_{k}^{j}\right\} _{k = 1}^{K_{j}}$. 
\end{itemize}

\subparagraph{Variables of assignment group to barycenter}
\begin{itemize}
\item Global transportation plan $\ba = (a_{jm}) \in \Pi(\frac{1}{J}\boldsymbol{1}_{J}, \bb)$ between $\locmeas$ and $\glomeas$.
\end{itemize}

\subparagraph{Variables of global clustering structures}
\begin{itemize}
\item Partial global transportation plans between local measure $P_{\bomega_{K_{j}}^{j}, \bTheta_{K_{j}}^{j}}$ and global measure $Q_{\bw^{m},\Psi_{k}^{m}}^{m}$: $\btau^{j,m} = \left\{ \tau_{kl}^{j,m}\right\} _{k,l}$ where
$\sum_{l}\tau_{kl}^{j,m}=\omega_{k}^{j}$ and $\sum_{l} \tau_{kl}^{j,m}=w_{l}^{m}$
for all $1 \leq j \leq J$ and $1 \leq m \leq \gloclus$.
\item Global atoms for global measure $\Psi_{L}^{m}=\left\{ \psi_{l}^{m}\right\} _{l=1}^{L}$.
and global mixing weights $\bw_{L}^{m}=\left\{ w_{l}^{m}\right\} _{l=1}^{L}$where
$w_{l}^{m}=\sum_{k}\tau_{kl}^{jm}$ for any $j$.
\end{itemize}
\subsection{Local clustering updates}
As being mentioned in the main text, to obtain updates for local weights $\bomega_{K_{j}}^{j}$ and local atoms $\bTheta_{K_{j}}^{j}$, we solve the following regularized composite transportation barycenter problem
\begin{align}
& \inf \limits_{\bomega_{K_{j}}^{j}, \bTheta_{K_{j}}^{j}} \compoop\left(P_{n_{j}}^{j}, P_{\bomega_{K_{j}}^{j},\bTheta_{K_{j}}^{j}}^{j}\right) - \lambda_{l} \entro(\bpi^{j}) + \sum_{m = 1}^{\gloclus} a_{jm}\compoop(P_{\bomega_{K_{j}}^{j},\bTheta_{K_{j}}^{j}}^{j}, Q_{\bw_{L}^{m},\Psi_{L}^{m}}^{m}) - \lambda_{g}\sum_{m=1}^{K} \entro(\btau^{j,m}). \nonumber
\end{align}
The above objective function can be rewritten as
\begin{align}
& \inf \limits_{\bomega_{K_{j}}^{j}, \bTheta_{K_{j}}^{j}} \inf_{\bpi^{j} \in \Pi(\frac{1}{n_{j}}\boldsymbol{1}_{n_{j}}, \bomega_{K_{j}}^{j})} \left\langle \bpi^{j},\bgamma^{j} \right\rangle - \lambda_{l} \entro(\bpi^{j}) + \sum_{m = 1}^{\gloclus} a_{jm} \inf \limits_{\btau^{j,m} \in \Pi(\bomega_{K_{j}}^{j}, \bw_{L}^{m})} \left\langle \btau^{j,m},\bglo^{j,m} \right\rangle \nonumber \\
& \hspace{24 em} - \lambda_{g}\sum_{m=1}^{K} \entro(\btau^{j,m}) \label{eq:local_clustering_updates_appendix}
\end{align}
where $\bgamma^{j}$ is the cost matrix between $P_{n_{j}}^{j}$ and $P_{\bomega_{K_{j}}^{j},\bTheta_{K_{j}}^{j}}^{j}$ that has a formulation as\begin{align}
[M^{j}]_{uv} = - \log f(X_{j,u}|\theta_{v}^{j}), \nonumber
\end{align} for $1 \leq u \leq n_{j}$ and $1 \leq v \leq K_{j}$. Additionally, the cost matrix $\bglo^{j,m}$ has the following formulation \begin{align}\gamma_{kl}^{j,m} & = A\left(\theta_{k}^{j}\right) - A\left(\psi_{l}^{m} \right) \nonumber  - \left\langle \grad A\left(\psi_{l}^{m}\right),\left(\theta_{k}^{j} - \psi_{l}^{m} \right)\right\rangle. \nonumber
\end{align}
\paragraph{Update local weights:} The idea for obtaining the local solutions of above objective function is similar to that in Section~\ref{Section:finite_mixture_composite}. 
\paragraph{Update local atoms:}
Given the updates for local weight $\bomega_{K_{j}}^{j}$, to obtain the update equation
for local atoms $\bTheta_{K_{j}}^{j}$, we optimize the following
objective function 
\begin{align}
\min_{\bTheta_{K_{j}}^{j}}-\sum_{u=1}^{n_{j}}\sum_{v = 1}^{K_{j}} & \pi_{uv}^{j}\log f(X_{j,u}|\theta_{v}^{j})+\sum_{m = 1}^{\gloclus} a_{jm}\sum_{v,l}\tau_{vl}^{j,m}\gamma_{kl}^{j,m}.\label{eqn:local_atom_update_multilevel}
\end{align}
Since $f(x|\theta)$ is an exponential family distribution with natural
parameter $\theta$, we can represent it as 
\begin{align*}
f(x|\theta) & = h(x) \exp\left(\left\langle T\left(x\right),\theta\right\rangle -A\left(\theta\right)\right),
\end{align*}
where $A\left(\theta\right)$ is the log-partition function which
is convex. Given that formulation of $f(x|\theta)$, our objective
function~(\ref{eqn:atom_update_mixture}) is equivalent to minimize
the following objective function 
\begin{align*}
\mathcal{F}_{\text{local}} & = - \sum_{u=1}^{n_{j}}\sum_{v = 1}^{K_{j}} \pi_{uv}^{j} \left(\left\langle T\left(X_{j,u}\right),\theta_{v}^{j}\right\rangle - A\left(\theta_{v}^{j}\right)\right) + \sum_{m = 1}^{\gloclus} a_{jm}\sum_{v,l}\tau_{kl}^{j,m}\biggr[A\left(\theta_{v}^{j}\right) - A\left(\psi_{l}^{m} \right) \nonumber \\
& \hspace{ 24 em} -\left\langle \grad A\left(\psi_{l}^{m}\right),\left(\theta_{v}^{j} - \psi_{l}^{m} \right)\right\rangle\biggr].
\end{align*}
By direct computation, $\mathcal{F}_{\text{local}}$ has the following partial derivative
with respect to $\theta_{v}^{j}$
\begin{align*}
\partdev{\mathcal{F}_{\text{local}}}{\theta_{v}^{j}} & = - \sum_{u = 1}^{n_{j}} \pi_{uv}^{j} \left(T(X_{j,u})-\grad A\left(\theta_{v}^{j}\right)\right) + \sum_{m = 1}^{\gloclus} a_{jm} \sum_{l = 1}^{L} \tau_{vl}^{j,m}\left[\grad A\left(\theta_{v}^{j}\right)-\grad A\left(\psi_{l}^{m}\right)\right] \\
 & = -\sum_{u = 1}^{n_{j}} \pi_{uv}^{j} T(X_{j,u}) + \omega_{v}^{j}\grad A\left(\theta_{v}^{j}\right) + \sum_{m = 1}^{C} a_{jm}\sum_{l = 1}^{L} \tau_{vl}^{j,m}\left(\grad A\left(\theta_{v}^{j}\right)-\grad A\left(\psi_{l}^{m}\right)\right),
\end{align*}
where in the last equality,
we use the identity $\sum_{u = 1}^{n_{j}}\pi_{uv}^{j} = \omega_{v}^{j}$.

Given the above partial derivatives, we can update the atoms $\theta_{v}^{j}$ to be the solution of the following
equation 
\begin{align}
\grad A\left(\theta_{v}^{j}\right) & =\frac{\sum \limits_{m = 1}^{C} a_{jm}\sum \limits_{l = 1}^{L} \tau_{vl}^{j,m} \grad A\left(\psi_{l}^{m}\right) + \sum \limits_{u = 1}^{n_{j}} \pi_{uv}^{j} T(X_{j,u})}{\sum \limits_{m = 1}^{C} a_{jm}\sum \limits_{l = 1}^{L} \tau_{vl}^{j,m} + \omega_{v}^{j}} \label{eqn:local_atoms_update_formulation}
\end{align}
\subsection{Computing global transportation plan}
Given the updates for local weights $\bomega_{K_{j}}^{j}$ and local atoms $\bTheta_{K_{j}}^{j}$ for $1 \leq j \leq J$, we now develop an update on for global transportation plan $\ba = (a_{jm})$ between $\locmeas$ and $\glomeas$. Our strategy for the update relies on solving the following objective function
\begin{align*}
 & \inf\limits _{\ba}\sum_{j,m}a_{jm}\compoop\left(P_{\bomega_{K_{j}}^{j},\bTheta_{K_{j}}^{j}}^{j},Q_{\bw_{L}^{m},\bPsi_{L}^{m}}^{m}\right)-\lambda_{a} \entro\left(\ba\right).
\end{align*}
where $\ba$ in the above infimum satisfies the constraint $\ba\boldsymbol{1}_{K}=\frac{1}{J}\boldsymbol{1}_{n}$. By means of Lagrangian multiplier, the above objective function can be rewritten as
\begin{align*}
\mathcal{F}_{\text{global}} & =\sum_{j,m}a_{jm}\compoop\left(P_{\bomega_{K_{j}}^{j},\bTheta_{K_{j}}^{j}}^{j},Q_{\bw_{L}^{m},\bPsi_{L}^{m}}^{m}\right) - \lambda_{a}\entro\left(\ba\right)+\kappa_{a}\sum_{j = 1}^{J} \left(\sum_{m = 1}^{\gloclus} a_{jm}-\frac{1}{J}\right),
\end{align*}
The function $\mathcal{F}_{\text{global}}$ has the partial derivative with respect to $a_{jm}$ as follows
\begin{align*}
\partdev{\mathcal{F}_{\text{global}}}{a_{jm}} & = \compoop\left(P_{\bomega_{K_{j}}^{j},\bTheta_{K_{j}}^{j}}^{j},Q_{\bw_{L}^{m},\bPsi_{L}^{m}}^{m}\right)+\lambda_{a}\log a_{jm}+\kappa_{a}\\
 & = \sum_{k,l}\tau_{kl}^{j,m}\gamma_{kl}^{j,m}+\lambda_{a}\ln a_{jm}+\kappa_{a}.
\end{align*}
Setting the above derivative to 0 and invoking the constraint that $\sum_{m =1}^{\gloclus} a_{jm}=\frac{1}{J}$,
we find that
\begin{align}
a_{jm} & =\frac{1}{J}\frac{\exp\left(-\sum_{k,l}\tau_{kl}^{j,m}\gamma_{kl}^{j,m}\right)^{1/\lambda_{a}}\,}{\sum_{m}\exp\left(-\sum_{k,l}\tau_{kl}^{j,m}\gamma_{kl}^{j,m}\right)^{1/\lambda_{a}}} \label{eqn:global_transport_update_formulation}
\end{align}
for $1 \leq j \leq J$ and $1 \leq m \leq L$.
\subsection{Global clustering updates}
Given the updates with local weights and atoms as well as the global transportation plan, we are now ready to develop an update for global weights $\bw_{L}^{m}$ and global atoms $\bPsi_{L}^{m}$ for $1 \leq m \leq \gloclus$. In particular, the objective
function for updating these global parameters are as follows
\begin{align*}
& \min_{\{\bw_{L}^{m}\}, \{\bPsi_{L}^{m}\}} \sum_{j = 1}^{J} \sum_{m = 1}^{\gloclus} a_{jm}\compoop(P_{\bomega_{K_{j}}^{j},\bTheta_{K_{j}}^{j}}^{j}, Q_{\bw_{L}^{m},\Psi_{L}^{m}}^{m}) - \lambda_{g}\sum_{j = 1}^{J} \sum_{m = 1}^{\gloclus} \entro(\btau^{j,m})
\end{align*}
The above objective function can be rewritten as
\begin{align*}
 & \min \limits_{\{\bw_{L}^{m}\},\{\boldsymbol{\Psi}_{L}^{m}\}}\sum_{j=1}^{J} \sum_{m = 1}^{\gloclus} a_{jm} \inf \limits_{\btau^{j,m} \in \Pi(\bomega_{K_{j}}^{j}, \bw_{L}^{m})} \left\langle \btau^{j,m},\bglo^{j,m} \right\rangle + \lambda_{g}\sum_{k,l}\tau_{kl}^{j,m}\left(\log\tau_{kl}^{j,m}-1\right)\biggr)
\end{align*}
Given the above objective function, for each $m$, to update the global weights $\bw_{L}^{m}$ and global atoms $\bPsi_{L}^{m}$, we consider the following composite transportation barycenter
\begin{align*}
 & \min \limits_{\bw_{L}^{m},\boldsymbol{\Psi}_{L}^{m}}\sum_{j=1}^{J} a_{jm} \inf \limits_{\btau^{j,m} \in \Pi(\bomega_{K_{j}}^{j}, \bw_{L}^{m})} \left\langle \btau^{j,m},\bglo^{j,m} \right\rangle + \lambda_{g}\sum_{k,l}\tau_{kl}^{j,m}\left(\log\tau_{kl}^{j,m}-1\right)\biggr).
\end{align*}
\paragraph{Update global weights:}
Given the above objective function, the idea for updating the global weights $\bw_{L}^{m}$ is similar to Algorithm 1 in~\cite{Cuturi-2014}.
\paragraph{Update partial transportation plans:}
Once global weights are obtained, we can use Algorithm 3 in~\cite{Cuturi-2014} to update the optimal partial transportation plans $\btau^{j,m}$ between local measure  $P_{\bomega_{K_{j}}^{j}, \bTheta_{K_{j}}^{j}}$ and global measure $Q_{\bw^{m},\Psi_{k}^{m}}^{m}$.
\paragraph{Update global atoms:}
With the updates for the global weight $\bw_{L}^{m}$, to obtain the update equation
for global atoms $\bPsi_{L}^{m}$, we minimize the following
objective function
\begin{align*}
\mathcal{F}_{\text{p-global}} & =\sum_{j = 1}^{J} a_{jm}\sum_{k,l}\tau_{kl}^{j,m}\gamma_{kl}^{j,m}\\
 & =\sum_{j = 1}^{J} a_{jm} \sum_{k,l} \tau_{kl}^{j,m}\biggr(A\left(\theta_{k}^{j}\right) - A\left(\psi_{l}^{m} \right) - \left\langle \grad A\left(\psi_{l}^{m} \right), \theta_{k}^{j}-\psi_{l}^{m} \right\rangle \biggr).
\end{align*}
Taking the derivative of $\mathcal{F}_{\text{p-global}}$ with respect to $\psi_{l}^{m}$ and setting it to zero, we find that
\begin{align*}
\partialdev{\mathcal{F}_{\text{p-global}}}{\psi_{l}^{m}} & =\sum_{j=1}^{J}a_{jm}\sum_{k}\tau_{kl}^{j,m}\grad^{2}A\left(\psi_{l}^{m}\right)\left(\psi_{l}^{m} - \theta_{k}^{j}\right) = 0.
\end{align*}
Since the log-partition function $A\left(\cdot\right)$ is convex,
$\grad^{2}A\left(\psi_{v}\right)$ is a positive-semidefinite matrix.
Therefore, we can choose $\sum_{j=1}^{J}a_{jm}\sum_{k}\tau_{kl}^{j,m}\left(\psi_{l}^{m} - \theta_{k}^{j}\right)=0$,
which means that
\begin{align*}
\psi_{l}^{m} & =\frac{\sum_{j=1}^{J}\sum_{k=1}^{K_{j}}a_{jm}\tau_{kl}^{j,m}\theta_{k}^{j}}{\sum_{j=1}^{J}\sum_{k=1}^{K_{j}}a_{jm}\tau_{kl}^{j,m}}. \label{eqn:global_atoms_update_formulation}
\end{align*}
\subsection{Proof for local convergence of Algorithm~\ref{algorithm:multilevel_composite_transportation_updates}}
Equipped with the above updates with local and global parameters of regularized MCT, we are ready to demonstrate the convergence of Algorithm~\ref{algorithm:multilevel_composite_transportation_updates} to local solution of objective function~\eqref{eq:regu_multi_compo_optimal} of regularized MCT in Theorem~\ref{theorem:local_convergence_multilevel_composite}. To simplify the argument, we only provide proof sketch for this theorem.

In particular, we denote $\bomega_{K_{j}}^{j,(t)}$ and
$\bTheta_{K_{j}}^{j,(t)}$ as the updates of local weights and
local atoms in step $t$ of Algorithm~\ref{algorithm:multilevel_composite_transportation_updates}
for $t \geq 0$. Similarly, we denote $\bw_{L}^{m,(t)}$ and $\bPsi_{L}^{m,(t)}$ as the updates of global weights and global atoms at step $t$. Furthermore, we denote 
\begin{align*}
g(\{\bomega_{K_{j}}^{j,(t)}\},\{\bTheta_{K_{j}}^{j,(t)}\},\{\bw_{L}^{m,(t)}\},\{\bPsi_{L}^{m,(t)}\})
& : = \sum_{j=1}^{J} \compoop\left(P_{n_{j}}^{j}, P_{\bomega_{K_{j}}^{j},\bTheta_{K_{j}}^{j}}^{j}\right) + \zeta \compoop\left(\locmeas, \glomeas\right)\nonumber \\
& \hspace{14 em} - R\left(\bpi,\btau,\ba\right).
\end{align*}
Then, according to local clustering updates step, we would have 
\begin{align*}
g(\{\bomega_{K_{j}}^{j,(t)}\},\{\bTheta_{K_{j}}^{j,(t)}\},\{\bw_{L}^{m,(t)}\},\{\bPsi_{L}^{m,(t)}\}) \geq g(\{\bomega_{K_{j}}^{j,(t+1)}\},\{\bTheta_{K_{j}}^{j,(t+1)}\},\{\bw_{L}^{m,(t)}\},\{\bPsi_{L}^{m,(t)}\}).
\end{align*}
On the other hand, invoking the global clustering updates step, we achieve
\begin{align*}
g(\{\bomega_{K_{j}}^{j,(t+1)}\},\{\bTheta_{K_{j}}^{j,(t+1)}\},\{\bw_{L}^{m,(t)}\},\{\bPsi_{L}^{m,(t)}\}) \geq g(\{\bomega_{K_{j}}^{j,(t+1)}\},\{\bTheta_{K_{j}}^{j,(t+1)}\},\{\bw_{L}^{m,(t+1)}\},\{\bPsi_{L}^{m,(t+1)}\})
\end{align*}
Governed by the above results, for any $t \geq 0$, the following holds
\begin{align*}
g(\{\bomega_{K_{j}}^{j,(t)}\},\{\bTheta_{K_{j}}^{j,(t)}\},\{\bw_{L}^{m,(t)}\},\{\bPsi_{L}^{m,(t)}\}) \geq g(\{\bomega_{K_{j}}^{j,(t+1)}\},\{\bTheta_{K_{j}}^{j,(t+1)}\},\{\bw_{L}^{m,(t+1)}\},\{\bPsi_{L}^{m,(t+1)}\}).
\end{align*}
As a consequence, we achieve the conclusion of Theorem~\ref{theorem:local_convergence_multilevel_composite}.
\end{document}